\pdfoutput=1

\documentclass[11pt]{article}

\usepackage[final]{acl}

\usepackage{times}
\usepackage{latexsym}

\usepackage{algorithm}
\usepackage{algorithmic}
\usepackage{graphicx}
\usepackage{booktabs}
\usepackage{algorithm}
\usepackage{algorithmic}
\usepackage{amsfonts,amssymb}
\usepackage{amsmath,bm}
\usepackage{subcaption}
\usepackage{diagbox}
\usepackage{url}
\usepackage{multirow} 
\usepackage{multicol} 
\usepackage{xcolor}
\usepackage{colortbl}
\usepackage{mathrsfs}
\usepackage{tabularx}
\usepackage{makecell}
\usepackage{amsfonts}
\usepackage{bbm}
\usepackage{arydshln}
\usepackage{booktabs}
\usepackage{bbding}
\usepackage{pifont}

\usepackage[T1]{fontenc}

\usepackage[utf8]{inputenc}

\usepackage{microtype}

\usepackage{inconsolata}

\title{Multi-modal Stance Detection: New Datasets and Model}

\author{Bin Liang$^{1,3,4\ast}$, Ang Li$^{1,3}$\thanks{\quad The first two authors contribute equally to this work.}, Jingqian Zhao$^{1,3}$, Lin Gui$^{5}$, Min Yang$^{6}$, \\ \bf Yue Yu$^{2}$, Kam-Fai Wong$^{4}$, and Ruifeng Xu$^{1,2,3}$\thanks{\quad Corresponding Author}\\
    $^{1}$ Harbin Institute of Technology, Shenzhen, China \\
    $^{2}$ Peng Cheng Laboratory, Shenzhen, China \\
    $^{3}$ Guangdong Provincial Key Laboratory of Novel Security Intelligence Technologies \\
    $^{4}$ The Chinese University of Hong Kong, Hong Kong, China~
    $^{5}$ King’s College London, UK \\
    $^{6}$ SIAT, Chinese Academy of Sciences, Shenzhen, China \\
    \texttt{bin.liang@cuhk.edu.hk},
    ~\texttt{angli@stu.hit.edu.cn},
    ~\texttt{xuruifeng@hit.edu.cn} \\
}

\begin{document}
\maketitle
\begin{abstract}
Stance detection is a challenging task that aims to identify public opinion from social media platforms with respect to specific targets. Previous work on stance detection largely focused on pure texts. In this paper, we study multi-modal stance detection for tweets consisting of texts and images, which are prevalent in today’s fast-growing social media platforms where people often post multi-modal messages. To this end, we create five new multi-modal stance detection datasets of different domains based on Twitter, in which each example consists of a text and an image. In addition, we propose a simple yet effective Targeted Multi-modal Prompt Tuning framework (\texttt{TMPT}), where target information is leveraged to learn multi-modal stance features from textual and visual modalities. Experimental results on our five benchmark datasets show that the proposed \texttt{TMPT} achieves state-of-the-art performance in multi-modal stance detection.
\end{abstract}

\section{Introduction}
Stance detection is an important task for learning public opinion from social media platforms, which aims to determine people's opinionated standpoint or attitude (\textit{e.g.}, {\em Favor}, {\em Against}, or {\em Neutral}, etc.) expressed in the content towards a specific target, topic, or proposition~\cite{somasundaran2010recognizing,augenstein-etal-2016-stance}. 
Existing conventional machine learning-based methods~\cite{hasan-ng-2013-stance,mohammad-etal-2016-semeval,ebrahimi-etal-2016-weakly} and deep learning-based methods~\cite{augenstein-etal-2016-stance,sun-etal-2018-stance,zhang-etal-2020-enhancing-cross,chen2021integrating,allaway-etal-2021-adversarial,liang2022zero} have made promising progress in different types of stance detection tasks for pure texts. 

\begin{figure}[!t]
\centering  
\includegraphics[width=0.95\linewidth]{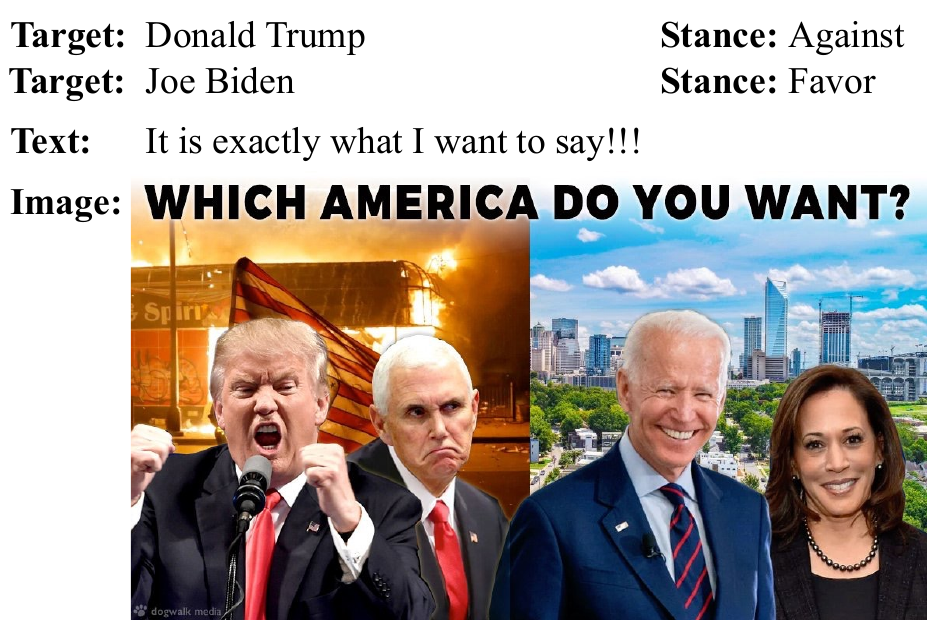}
\caption{An example of a user expressing an ``Against'' stance towards ``Donald Trump'' and a ``Favor'' stance towards ``Joe Biden'' using multi-modal information.}
\label{fig-example}
\end{figure}

However, more and more present-day social media platforms like Twitter allow people to post multi-modal messages, which encourages people to express their stances and opinions through multi-modal content, posting texts with images for example. 
That is, detecting stance from the pure text modality may not accurately identify the user's real view of a target. For example, Figure~\ref{fig-example} shows a post composed of a text and an image. The stance expression towards ``Donald Trump'' and ``Joe Biden'' in this example can not be accurately identified based on text information unless combined with the information of the visual modality. Therefore, how to detect users' stances on a topic from multi-modal posts might help better identify public opinion in social media.

For multi-modal stance identification, \citet{weinzierl-harabagiu-2023-identification} discussed the multi-modal stance towards frames of communication. Different from conventional stance detection tasks that concentrate on the stance towards several predefined targets, their focus was centered on the frames of communication within multi-modal posts. 
Therefore, aiming to push forward the research of multi-modal stance detection, we create five new datasets, in which each example consists of a target, a text, and an image. These datasets contain a total of 17,544 examples across 5 domains and 12 targets, including hot topics, politicians, and debates.

To deal with multi-modal stance detection, we propose a simple yet effective Targeted Multi-modal Prompt Tuning framework (\texttt{TMPT}), where the targeted prompt tuning is employed to adapt pre-trained models for learning stance features from different modalities.
Specifically, to leverage the target information in stance detection, we first devise targeted prompts for both textual and visual modalities. Then, the targeted prompts are fed to the pre-trained language model and pre-trained visual model to learn stance features for the target from different modalities. Further, a simple vector concatenation is used to fuse the features from different modalities for multi-modal stance detection.

The main contributions of our work are summarized as follows:

1) We manually annotate five new multi-modal stance detection datasets based on Twitter data from different domains. The release of the datasets would push forward the research in this field.

2) A simple yet effective targeted multi-modal prompting tuning framework is proposed to deal with multi-modal stance detection, where the target information is used to prompt the pre-trained models for learning multi-modal stance features.

3) A series of experiments on our datasets show that the proposed method significantly outperforms the baseline models\footnote{To facilitate future research, our datasets and code are publicly available at \url{https://github.com/Leon-Francis/Multi-Modal-Stance-Detection}}.

\begin{table*}[!t]
    \small
    \centering
    \setlength{\tabcolsep}{2.5pt}
    \renewcommand{\arraystretch}{1.1}
    \begin{tabular}{c|cccccccccc}
        \hline
        Dataset                                       & Target            & \multicolumn{9}{c}{\# Samples and Proportion of Labels}                                                    \\
        \hline
        \multicolumn{1}{c|}{\multirow{3}{*}{\textsc{Mtse}}} &                    & Favor &     \% & Against  &     \% & Neutral &     \% & -         & -      & Total \\ \cline{3-11}
                                                      & Donald Trump (DT)             & 231     &  14.03 & 1297    &  78.75 & 119     &   7.23 & -         & -      & 1647  \\
                                                      & Joe Biden (JB)             & 602     &  47.78 & 524     &  41.59 & 134     &  10.63 & -         & -      & 1260  \\
        \hline
        \multicolumn{1}{c|}{\multirow{2}{*}{\textsc{Mccq}}}     &                    & Favor   &     \% & Against &     \% & Neutral &     \% & -         & -      & Total \\ \cline{3-11}
                                                      & Chloroquine (CQ)       & 455     &  33.58 & 503     &  37.12 & 397     &  29.30 & -         & -      & 1355  \\
        \hline
        \multicolumn{1}{c|}{\multirow{6}{*}{\textsc{Mwtwt}}}    &                    & Support &     \% & Refute  &     \% & Comment &     \% & Unrelated &     \% & Total \\ \cline{3-11}
                                                      & CVS\_AET           & 426     &  24.38 & 65      &   3.72 & 866     &  49.57 & 390       &  22.32 & 1747  \\
                                                      & CI\_ESRX           & 321     &  35.71 & 91      &  10.12 & 298     &  33.15 & 189       &  21.02 & 899   \\
                                                      & ANTM\_CI           & 59      &   5.01 & 238     &  20.22 & 306     &  26.00 & 574       &  48.77 & 1177  \\
                                                      & AET\_HUM           & 94      &   9.82 & 287     &  29.99 & 267     &  27.90 & 309       &  32.29 & 957   \\
                                                      & DIS\_FOXA          & 417     &  13.97 & 103      &   3.45 & 1504    &  50.37 & 962       &  32.22 & 2986  \\
        \hline
        \multicolumn{1}{c|}{\multirow{3}{*}{\textsc{Mruc}}} &                    & Support &     \% & Oppose  &     \% & Neutral &     \% & -         & -      & Total \\ \cline{3-11}
                                                      & Russia (RUS)             & 16     &  1.4 & 763    &  68.74 & 331     &   29.82 & -         & -      & 1110  \\
                                                      & Ukraine (UKR)             & 742     &  68.64 & 39     &  3.61 & 300     &  27.75 & -         & -      & 1081  \\
        \hline
        \multicolumn{1}{c|}{\multirow{3}{*}{\textsc{Mtwq}}} &                    & Support &     \% & Oppose  &     \% & Neutral &     \% & -         & -      & Total \\ \cline{3-11}
                                                      & Mainland of China (MOC)             & 339     &  24.27 & 834    &  59.70 & 224     &   16.03 & -         & -      & 1397  \\
                                                      & Taiwan of China (TOC)             & 1595     &  82.73 & 74     &  3.84 & 259     &  13.43 & -         & -      & 1928  \\
        \hline
    \end{tabular}
    \caption{Label distribution of the five multi-modal stance detection datasets.}
    \label{label-distribution}
\end{table*}

\section{Related Work}
\paragraph{Textual Stance Detection}Various methods based on conventional machine learning~\cite{hasan-ng-2014-taking,mohammad-etal-2016-semeval} and deep learning~\cite{sun-etal-2018-stance,DBLP:conf/emnlp/ZhengSYX22,DBLP:conf/acl/LiC23,DBLP:conf/emnlp/LiLZZYX23} have been proposed to deal with the stance detection regarding a specific target. For performing stance detection in real-world scenarios, many existing methods focus on the task of zero-shot stance detection\footnote{Following~\cite{allaway2020zero}, the term "zero-shot" here refers to the model's ability to detect stance towards targets it has not encountered during training.}\cite{liu-etal-2021-enhancing,liang2022jointcl,wen-hauptmann-2023-zero,DBLP:conf/acl/ZhaoLC23}.

\paragraph{Prompt Tuning}Prompting~\cite{Liu-etal-2021-Pre-train} was originally aimed to design language instructions for pre-trained language models (PLMs) to transfer learning in downstream tasks~\cite{shin-etal-2020-AutoPrompt, Jiang-etal-2020-How}. Recent works begin to treat prompts as continuous vectors and optimize them during fine-tuning, called Prompt Tuning~\cite{li-liang-2021-prefix, Lester-etal-2021-The-Power, Liu-etal-2021-P-Tuning}.
Besides, \citet{radford2021learning, zhou-etal-2022-Learning, Ju-etal-2022-Prompting,jia2022vpt} introduce prompt into vision-language models to leverage the ability of prompt tuning in multi-modal tasks.

\begin{table}[t]
\small
\centering
\setlength{\tabcolsep}{2.5pt}
\renewcommand{\arraystretch}{1.1}
\begin{tabular}{l|c|ccccc}
\hline
      & Image & Person & Events & Words & Memes & Mixed \\
\hline
\textsc{Mtse}  & 43.7  & 32.7   & 8.2    & 24.0  & 25.0  & 10.1  \\
\textsc{Mccq}  & 44.1  & 12.2   & 22.9   & 26.7  & 16.3  & 21.9  \\
\textsc{Mwtwt} & 46.2  & 26.4   & 16.8   & 38.4  & 5.9   & 12.5  \\
\textsc{Mruc}  & 54.8  & 32.1   & 20.7   & 20.4  & 18.5  & 8.3   \\
\textsc{Mtwq}  & 41.0  & 36.8   & 42.9   & 8.1   & 2.4   & 9.8   \\
\hline
\end{tabular}
\caption{The statistics of whether the image conveys stance information (\%Image) and the type of each image (\%Person, \%Events, \%Words, \%Memes, \%Mixed).}
\label{tab:image_statistic}
\end{table}

\section{Multi-modal Stance Detection Datasets}  
\label{data_details}

Based on three open-source textual stance detection datasets: Twitter Stance Election 2020~\cite{kawintiranon-singh-2021-knowledge}, COVID-CQ \cite{mutlu2020stance}, and Will-They-Won’t-They \cite{conforti-etal-2020-will}, and two hot topics in recent years: Russo-Ukrainian Conflict\footnote{\url{https://en.wikipedia.org/wiki/Russo-Ukrainian_War}} and Taiwan Question\footnote{\url{https://en.wikipedia.org/wiki/Cross-Strait_relations}}. We create five multi-modal stance detection datasets of different domains to provide available data for this task: Multi-modal Twitter Stance Election 2020 (\textsc{Mtse}), Multi-modal COVID-CQ (\textsc{Mccq}), Multi-modal Will-They-Won’t-They (\textsc{Mwtwt}), Multi-modal Russo-Ukrainian Conflict (\textsc{Mruc}) and Multi-modal Taiwan Question (\textsc{Mtwq}).

\subsection{Data Collection}

We use Twitter Streaming API\footnote{\url{https://developer.twitter.com/en/products/twitter-api}} to  collect tweets with corresponding keywords of five datasets (shown in Appendix~\ref{sec:keywords}), keeping the posts containing text in English and at least one image or video/GIF. For videos/GIFs, we retain their first frame because the visual information contained in consecutive frames may be very similar.
For posts with multiple images, we combine the text with each image to form multiple samples due to the fact that different images may contain completely different visual information.
Finally, we obtain 130000, 90000, 60000, 100000, and 80000 candidate examples of five datasets, respectively. 

\subsection{Data Annotation}
\label{data-anno}

To ensure consistency with previous stance detection work, we follow the guidelines of Twitter Stance Election 2020~\cite{kawintiranon-singh-2021-knowledge}, COVID-CQ \cite{mutlu2020stance}, and Will-They-Won't-They \cite{conforti-etal-2020-will} to annotate the multi-modal stance of \textsc{Mtse}, \textsc{Mccq} and \textsc{Mwtwt}. For \textsc{Mruc} and \textsc{Mtwq}, the annotation guidelines are shown in Appendix~\ref{sec:guidelines}.
The meaningless/noisy posts or those that do not comply with Twitter's policies or annotation guidelines are discarded during the annotation.
We invite eight experienced researchers\footnote{We recruit experienced researchers who have worked on multi-modal learning over 3 years.} to label the stance for each example. 
Each sample will be annotated by three different annotators, and the gold label is obtained by majority vote. For the disagreed results among the three annotators, we invited three additional annotators to annotate and then performed a majority vote to obtain the gold label\footnote{During the annotation process, only 2.54\% of the data need to be allocated to additional annotators.}. 

\subsection{Quality Assessment}

We use Cohen's Kappa Statistic to evaluate the inter-annotator agreement~\cite{cohen1960coefficient}.
The average Cohen's Kappa between our annotator pairs for \textsc{Mtse} is 0.703, for \textsc{Mccq} is 0.689, for \textsc{Mwtwt} is 0.729, for \textsc{Mruc} is 0.752, and for \textsc{Mtwq} is 0.691. This demonstrates that the Kappa scores of all datasets are substantial. In addition, the average Cohen's Kappa reported in the related textual stance detection dataset Will-They-Won't-They \cite{conforti-etal-2020-will} is 0.67, which also indicates the high quality of our new datasets from another angle.

\begin{figure*}[!t]
\centering  
\includegraphics[width=0.90\linewidth]{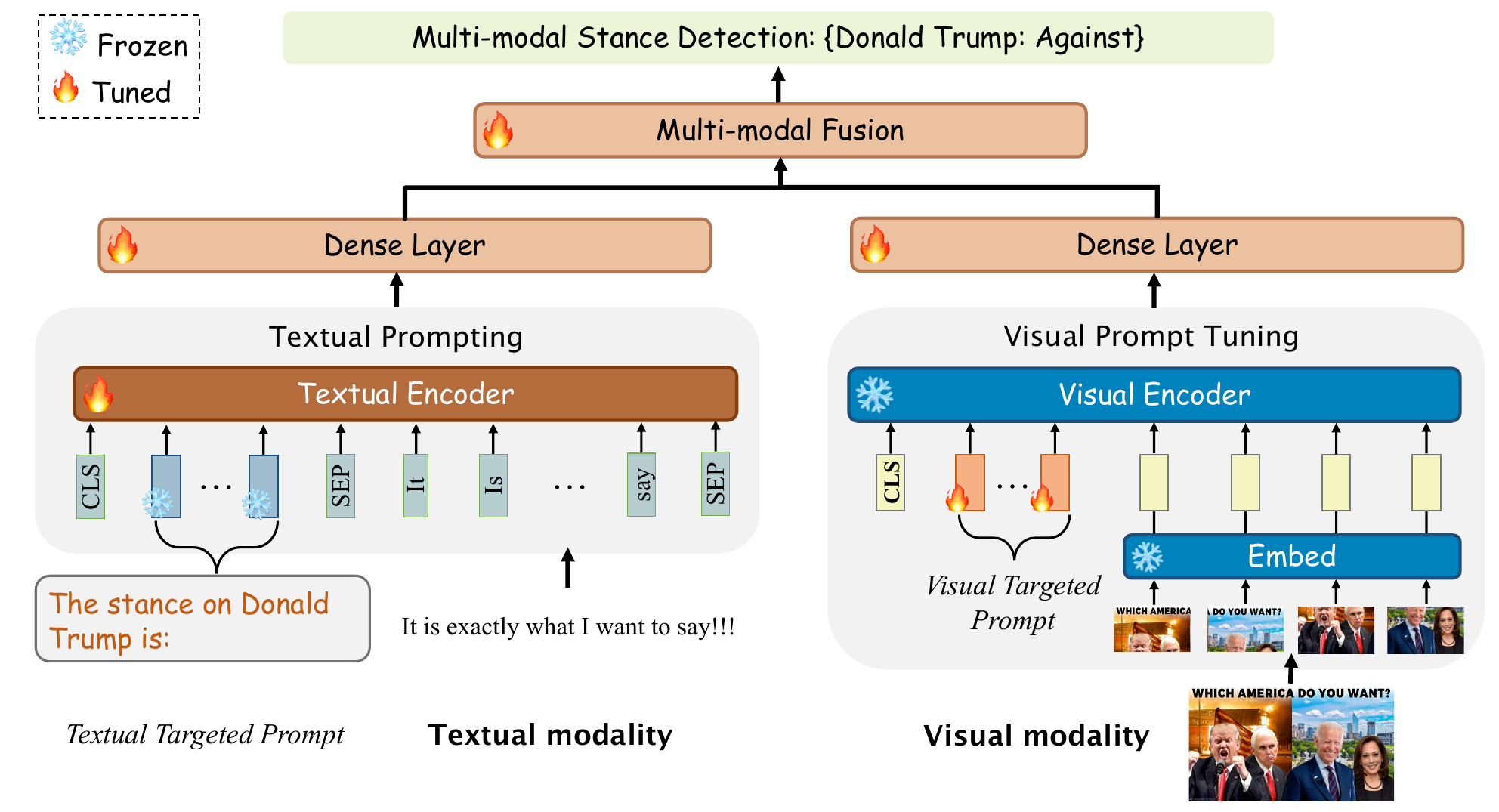}
\caption{The overall architecture of our proposed \texttt{TMPT}. Textual Prompting is devised for adapting the large pre-trained language model. Visual Prompt Tuning is devised for adapting the large pre-trained vision model.}
\label{fig-model}
\end{figure*}

\subsection{Data Analysis}
\label{data-analysis}
We finally got 17,544 well-annotated multi-modal samples across 5 domains and 12 targets. Each example consists of a text in English and an associated image. The statistics of datasets are reported in Table~\ref{label-distribution}. Note that differences in label distribution between targets are common, which is also observed in other textual stance detection datasets~\cite{mohammad-etal-2016-semeval,mutlu2020stance,conforti-etal-2020-will}.
Through analyzing the indication of whether the image conveys stance information introduced in Section~\ref{data-anno}, and analyzing the type of each image, we obtained statistics shown in Table~\ref{tab:image_statistic}. It can be seen that nearly half of the stance information in all five datasets comes from visual modalities. Further, the proportion of image types varies greatly among different datasets. This complicates image comprehension for stance detection, which is also the main challenge of multi-modal stance detection.

\section{Methodology}
In this section, we introduce our proposed targeted multi-modal prompt tuning framework (\texttt{TMPT}) in detail.
Given a text $S$ and an image $I$, the goal of multi-modal stance detection is to identify the stance label $y$ for the specific target $t$ based on $S$ and $I$. Therefore, to leverage the target information for multi-modal stance detection, we design targeted multi-modal prompt tuning for both textual and visual modalities, which are utilized to prompt the pre-trained models for learning multi-modal stance features. The architecture of our \texttt{TMPT} is illustrated in Figure~\ref{fig-model}, containing four main components: 1) \textit{Textual Prompt Tuning}, which encodes the input of textual modality based on the textual targeted prompt; 2) \textit{Visual Prompt Tuning}, which encodes the input of visual modality based on the visual targeted prompt; 3) \textit{Multi-modal Fusion}, which fuses the representations from textual and visual modalities to capture the stance features. 4) \textit{Multi-modal Stance Detection}, which derives the stance label for an input example according to the multi-modal stance features.

\begin{table}[!t]
    \small
    \centering
    \setlength{\tabcolsep}{1.0pt}
    \renewcommand{\arraystretch}{1.1}
    \begin{tabularx}{\linewidth}{c|c|X}
    \hline
    Dataset & Target & Textual Targeted Prompt \\
    \hline
    \multirow{1}*{\textsc{Mtse}} & \multirow{1}*{DT} & The~stance~on~Donald Trump~is: \\
    \hline
    \multirow{4}*{\textsc{Mccq}} & \multirow{4}*{CQ} & The stance on the use of Chloroquine and Hydroxychloroquine for the treatment or prevention from the coronavirus or COVID 19 is: \\
    \hline
    \multirow{2}*{\textsc{Mwtwt}} & \multirow{2}*{CVS\_AET} & The stance on merger and acquisition between CVS Health and Aetna is:  \\
    \hline
    \multirow{1}*{\textsc{Mruc}} & RUS & The stance on Russia is:  \\
    \hline\multirow{1}*{\textsc{Mtwq}} & MOC & The stance on Mainland of China is:  \\
    \hline
    \end{tabularx}
    \caption{The example of textual targeted prompts.}
    \label{tab:textual-prompt}
\end{table}

\subsection{Textual Prompt Tuning}
\paragraph{Textual Prompt Construction}Inspired by the textual prompt tuning~\cite{DBLP:journals/csur/LiuYFJHN23}, considering the characteristics of stance detection, we devise a textual targeted prompt for each text to adapt the pre-trained language model to the stance detection task. As the example shown in Figure~\ref{fig-model}, take the target ``Donald Trump'' as an example, the textual targeted prompt is defined as:
\begin{equation}
\begin{split}
\mathcal{P}^T=\mathrm{The~stance~on~Donald~Trump~is:}
\end{split}
\end{equation}

That is, the textual targeted prompts are designed according to the targets and the purpose of stance detection. The examples of textual targeted prompts are shown in Table~\ref{tab:textual-prompt}. Other textual prompt settings are introduced in Section~\ref{textual-prompt}.

\paragraph{Textual Encoder}Based on the textual targeted prompts, we construct the input of each text on target $t$. Given a text $S$ consists of a sequence of words $S=\{w_i\}_{i=1}^{n}$, $n$ is the length of $S$. The input of the textual modality is represented as:
\begin{equation}
\mathcal{I}^T=\texttt{[CLS]}\mathcal{P}^T\texttt{[SEP]}s\texttt{[SEP]}
\end{equation}

Then, we adopt the pre-trained uncased BERT-base model~\cite{devlin-etal-2019-bert} to map each textual token into a $d^T$-dimensional embedding:
\begin{equation}
[{\boldsymbol x}^T_{\texttt{[CLS]}},{\boldsymbol x}^T_1,\cdots,{\boldsymbol x}^T_{m+n+2}]={\rm BERT}(\mathcal{I}^T)
\end{equation}
Where ${\boldsymbol x}^T_{\texttt{[CLS]}}$ represents the embedding of the \texttt{[CLS]} token, $m$ is the length of the textual targeted prompt. In this way, the pre-trained language model can encode the input text according to the target, and obtain the feature representation that contains the target-specific stance information.

\subsection{Visual Prompt Tuning}
\paragraph{Visual Embedding Layer}Following~\cite{dosovitskiy2021an}, we first divide the image $I$ into $r$ fixed-sized patches $I=\{{\boldsymbol p}_i \in \mathbb {R}^{l^2\times c}\}_{i=1}^r$, where $(l,l)$ is the resolution of each patch, $c$ is the number of channels. Then, following~\cite{jia2022vpt}, we flatten  the patches and map to $d^V$-dimensional vector with a trainable linear embedding projection ${\boldsymbol E}$.  We refer to the output of this projection as the patch embeddings. Then we concatenate these patch embeddings in the sequence dimension and added the standard learnable 1D position embeddings ${\boldsymbol E}_{pos}$ to the patch embeddings to retain positional information:
\begin{gather}
{\boldsymbol v}_j={\boldsymbol p}_j{\boldsymbol E}\quad\quad {\boldsymbol v}_j \in \mathbb {R}^d, j\in \{1,r\} \\
{\boldsymbol V}^0=[\{{\boldsymbol v}_j\}_{j=1}^r]+{\boldsymbol E}_{pos}
\end{gather}
Where ${\boldsymbol V}^0$ is the embedding for the input image.

\paragraph{Visual Prompt Construction}Inspired by the visual prompt tuning proposed by~\cite{jia2022vpt}, we devise visual targeted prompt, aiming to instruct the pre-trained vision model to learn the features according to the specific target.
Specifically, we introduce continuous embedding (visual prompt tokens), as the visual prompt for target $t$.
Each prompt consists of $\lambda$ learnable embedding, which can be formulated as follows:
\begin{equation}
\mathcal{P}^V=\{{\boldsymbol e}_i \in \mathbb {R}^{d^V}| 1\leq i \leq \lambda\}
\end{equation}
Here, each different target corresponds to a different set of visual prompt embedding. In Section~\ref{ana-visual-prompting}, we introduced different initialization methods of visual prompt embedding.

\paragraph{Visual Encoder}Based on the visual embedding ${\boldsymbol V}^0$ and the visual targeted prompts $\mathcal{P}^V$, we use the pre-trained Vision Transformer model ViT~\cite{dosovitskiy2021an} with $N$ layers to encode the input image for learning visual stance features on target $t$. Here, the targeted prompts are inserted into the first Transformer layer $L^1$. Therefore, the first layer targeted prompted ViT is defined as:
\begin{equation}
[{\boldsymbol x}^V_{\texttt{[CLS]}_1},{\boldsymbol Z}_1,{\boldsymbol V}_1]=L^1([{\boldsymbol x}^V_{\texttt{[CLS]}_0},\mathcal{P}^V,{\boldsymbol V}_0])
\end{equation}

For layer $k \in \{2,N\}$, the targeted prompted ViT is defined as:
\begin{equation}
[{\boldsymbol x}^V_{\texttt{[CLS]}_k},{\boldsymbol Z}_k,{\boldsymbol V}_k]=L^k([{\boldsymbol x}^V_{\texttt{[CLS]}_{k-1}},{\boldsymbol Z}_{k-1},{\boldsymbol V}_{k-1}])
\end{equation}

Here, we use ${\boldsymbol x}^V_{\texttt{[CLS]}}$ to represent the embedding of \texttt{[CLS]} token learned by the final Transformer layer $N$, \textit{i.e.}, ${\boldsymbol x}^V_{\texttt{[CLS]}}={\boldsymbol x}^V_{\texttt{[CLS]}_N}$.
By using the visual prompts corresponding to each target, the pre-trained visual model can encode the input image according to the target, and obtain the feature representation that contains the target-specific stance information.

\subsection{Multi-modal Fusion}
Based on the embeddings of \texttt{[CLS]} tokens, we first use two dense layers with Leaky ReLU~\cite{maas2013rectifier} to derive the hidden representations of textual and visual modalities:
\begin{gather}
{\boldsymbol h}^T=f({\boldsymbol W}^T{\boldsymbol x}^T_{\texttt{[CLS]}}+{\boldsymbol b}^T)\\
{\boldsymbol h}^V=f({\boldsymbol W}^V{\boldsymbol x}^V_{\texttt{[CLS]}}+{\boldsymbol b}^V)
\end{gather}
Where ${\boldsymbol h}^T \in \mathbb {R}^{d^h}$ and ${\boldsymbol h}^V \in \mathbb {R}^{d^h}$ are the hidden representation of textual and visual modalities respectively.  $d^h$ is the dimensionality of the hidden representation. $f(\cdot)$ represents the Leaky ReLU activation function. ${\boldsymbol W}^T \in \mathbb {R}^{d^h\times d^T}$ and ${\boldsymbol W}^V \in \mathbb {R}^{d^h\times d^V}$ are weight matrices. ${\boldsymbol b}^T \in \mathbb {R}^{d^h}$ and ${\boldsymbol b}^V \in \mathbb {R}^{d^h}$ are biases.

In our \texttt{TMPT}, vector concatenation is used to fuse the feature vectors from different modalities:
\begin{equation}
{\boldsymbol h}={\boldsymbol h}^T \oplus {\boldsymbol h}^V
\end{equation}
Where ${\boldsymbol h} \in \mathbb {R}^{2d^h}$ is the final multi-modal stance representation. $\oplus$ represents the vector concatenation operation.

\subsection{Multi-modal Stance Detection}
Then, the final multi-modal stance representation is fed into a fully-connected layer with a softmax function to capture a probability distribution $\hat{\boldsymbol y}\in \mathbb {R} ^{d^p}$ in the stance decision space:
\begin{equation}
\hat{\boldsymbol y}={\rm{softmax}}({\boldsymbol W}^o{\boldsymbol h}+{\boldsymbol b}^o)
\end{equation}
Where ${d^p}$ is the dimensionality of stance labels. ${\boldsymbol W}^o\in \mathbb {R} ^{d^p\times2d^h}$ and ${\boldsymbol b}^o \in \mathbb {R} ^{d^p}$ are  weight matrix and bias respectively.

\begin{table}[t]
\small
\centering
\setlength{\tabcolsep}{2.5pt}
\renewcommand{\arraystretch}{1.1}
\begin{tabular}{cccccc}
\hline
Task                        & Dataset                & Target    & \# Train & \# Valid & \# Test \\
\hline
\multirow{12}{*}{In-target} & \multirow{2}{*}{MTSE}  & DT        & 1150     & 170      & 327     \\
                            &                        & JB        & 882      & 128      & 250     \\
                            \cdashline{2-6}[2pt/3pt]
                            & MCCQ                   & CQ        & 934      & 141      & 280     \\
                            \cdashline{2-6}[2pt/3pt]
                            & \multirow{5}{*}{MWTWT} & CSV\_AET  & 1216     & 179      & 352     \\
                            &                        & CI\_ESRX  & 628      & 91       & 180     \\
                            &                        & ANTM\_CI  & 825      & 114      & 238     \\
                            &                        & AET\_HUM  & 674      & 97       & 186     \\
                            &                        & DIS\_FOXA & 2081     & 306      & 599     \\
                            \cdashline{2-6}[2pt/3pt]
                            & \multirow{2}{*}{MRUC}  & RUS       & 777      & 111      & 222     \\
                            &                        & UKR       & 756      & 108      & 217     \\
                            \cdashline{2-6}[2pt/3pt]
                            & \multirow{2}{*}{MTWQ}  & MOC       & 977      & 140      & 280     \\
                            &                        & TOC       & 1349     & 193      & 386     \\
\hline
\multirow{10}{*}{Zero-shot} & \multirow{2}{*}{MTSE}  & DT        & 1114     & 146      & 1647    \\
                            &                        & JB        & 1434     & 212      & 1260    \\
                            \cdashline{2-6}[2pt/3pt]
                            & \multirow{4}{*}{MWTWT} & CVS\_AET  & 5253     & 737      & 1747    \\
                            &                        & CI\_ESRX  & 5994     & 841      & 899     \\
                            &                        & ANTM\_CI  & 5694     & 804      & 1177    \\
                            &                        & AET\_HUM  & 5884     & 840      & 957     \\
                            \cdashline{2-6}[2pt/3pt]
                            & \multirow{2}{*}{MRUC}  & RUS       & 945      & 136      & 1110    \\
                            &                        & UKR       & 971      & 139      & 1081    \\
                            \cdashline{2-6}[2pt/3pt]
                            & \multirow{2}{*}{MTWQ}  & MOC       & 1686     & 242      & 1397    \\
                            &                        & TOC       & 1222     & 175      & 1928    \\
\hline
\end{tabular}
\caption{Statistics of the experimental data.}
\label{dataset-statistic}
\end{table}

\begin{table*}[!t]
\small
\centering
\setlength{\tabcolsep}{2.5pt}
\renewcommand{\arraystretch}{1.1}
\setlength{\tabcolsep}{0.62mm}{
\begin{tabular}{llccccccccccccccccc}
\hline
\multirow{2}*{\textsc{Modality}}        & \multirow{2}*{\textsc{Method}}                & \multicolumn{2}{c}{MTSE}        &  & MCCQ           &  & \multicolumn{5}{c}{MWTWT}                                                          &  & \multicolumn{2}{c}{MRUC}     &  & \multicolumn{2}{c}{MTWQ} \\
\cline{3-4} \cline{6-6} \cline{8-12} \cline{14-15} \cline{17-18}
\multicolumn{1}{l}{}         &                        & DT             & JB             &  & CQ             &  & CA       & CE       & AC       & AH       & DF      &  & RUS            & UKR            &  & MOC             & TOC            \\
\hline
\multirow{5}{*}{Textual}     & BERT                & 48.25          & 52.04          &  & 66.57          &  & 75.62          & 60.85          & 63.05          & 59.24          & 81.53          &  & 41.25          & 46.80          &  & 57.77          & 45.91          \\
                             & RoBERTa                & 58.39          & 60.79          &  & 66.57          &  & 69.56          & 65.03          & 69.74          & 67.99          & 79.21          &  & 39.52          & 57.66          &  & 55.22          & 48.88          \\
                             & KEBERT                 & 64.50          & 69.81          &  & 66.84          &  & 71.67          & 67.56          & 69.29          & 69.74          & 80.57          &  & 41.55          & 59.01          &  & 58.15           & 47.75          \\
                             \cdashline{2-18}[2pt/3pt]
                             & LLaMA2                 & 53.23          & 52.67          &  & 47.40          &  & 34.89          & 41.95          & 49.09          & 44.32          & 30.21          &  & 38.84          & 38.54          &  & 55.31           & 46.51          \\
                             & GPT4                   & 68.74          & 66.39          &  & 65.84          &  & 63.14          & 65.12          & \textbf{69.93}          & \textbf{71.62}          & 52.69          &  & 41.64          & 53.76          &  & 58.05           & 49.81          \\
\hline
\multirow{3}{*}{Visual}      & ResNet                 & 37.89          & 38.59          &  & 47.16          &  & 39.89          & 42.20          & 43.52          & 37.05          & 50.34          &  & 35.10          & 40.00          &  & 42.02           & 33.94          \\
                             & ViT                    & 40.48          & 40.42          &  & 46.64          &  & 46.63          & 50.00          & 40.16          & 46.32          & 50.86          &  & 33.31          & 39.87          &  & 38.63           & 35.53          \\
                             & SwinT                  & 39.89          & 40.43          &  & 48.80          &  & 46.30          & 46.99          & 41.02          & 47.39          & 51.32          &  & 35.01          & 40.89          &  & 35.03           & 35.47          \\
\hline
\multirow{7}{*}{\begin{tabular}[c]{@{}c@{}}Multi-\\ modal\end{tabular}} & BERT+ViT               & 41.86          & 45.82          &  & 61.32          &  & 63.20          & 44.71          & 56.45          & 46.85          & 73.71          &  & 39.28          & 48.41          &  & 47.47           & 40.86          \\
                             & ViLT                   & 35.32          & 48.24          &  & 47.85          &  & 62.70          & 56.44          & 58.06          & 60.22          & 73.66          &  & 34.62          & 42.41          &  & 44.43           & 59.51          \\
                             & CLIP                   & 53.22          & 65.83          &  & 63.65          &  & 70.93          & 67.17          & 67.43          & 70.86          & 79.06          &  & 44.99          & 59.86          &  & 55.29           & 40.98          \\
                             \cdashline{2-18}[2pt/3pt]
                             & Qwen-VL                & 43.31          & 45.13          &  & 50.51          &  & 43.06          & 45.49          & 49.79          & 46.04          & 27.73          &  & 36.50          & 40.78          &  & 42.14           & 39.34          \\
                             & GPT4-Vision                 & \textbf{70.46}          & \textbf{72.82}          &  & 61.63          &  & 44.59          & 47.07          & 57.47          & 57.90          & 37.61          &  & 44.83          & 56.40          &  & 66.72           & 56.90          \\
\cdashline{2-18}[2pt/3pt]
& \cellcolor[gray]{0.92}\texttt{TMPT}       & \cellcolor[gray]{0.92}55.41           & \cellcolor[gray]{0.92}61.61           & \cellcolor[gray]{0.92} & \cellcolor[gray]{0.92}67.67           & \cellcolor[gray]{0.92} & \cellcolor[gray]{0.92}\textbf{76.60}           & \cellcolor[gray]{0.92}63.19           & \cellcolor[gray]{0.92}67.25           & \cellcolor[gray]{0.92}62.92           & \cellcolor[gray]{0.92}81.19  & \cellcolor[gray]{0.92} & \cellcolor[gray]{0.92}43.56           & \cellcolor[gray]{0.92}59.24           & \cellcolor[gray]{0.92} & \cellcolor[gray]{0.92}55.68           & \cellcolor[gray]{0.92}46.82           \\
& \cellcolor[gray]{0.92}\texttt{TMPT}+\texttt{CoT} & \cellcolor[gray]{0.92}66.61  & \cellcolor[gray]{0.92}68.75  & \cellcolor[gray]{0.92} & \cellcolor[gray]{0.92}\textbf{71.79}$^\star$  & \cellcolor[gray]{0.92} & \cellcolor[gray]{0.92}74.40  & \cellcolor[gray]{0.92}\textbf{69.96}$^\star$  & \cellcolor[gray]{0.92}68.43  & \cellcolor[gray]{0.92}63.00  & \cellcolor[gray]{0.92}\textbf{82.71}$^\star$           & \cellcolor[gray]{0.92} & \cellcolor[gray]{0.92}\textbf{45.04}$^\star$  & \cellcolor[gray]{0.92}\textbf{60.52}  & \cellcolor[gray]{0.92} & \cellcolor[gray]{0.92}\textbf{68.95}$^\star$  & \cellcolor[gray]{0.92}\textbf{59.87}$^\star$  \\
\hline
\end{tabular}
}
\caption{Experimental results (\%) of in-target multi-modal stance detection. The dark background results are for our \texttt{TMPT}. Best scores of each group are in bold. Results with $\star$ denote the significance tests of our \texttt{TMPT} over the baseline models at $p\rm{-}value<0.05$. The dashed line represents a separation between fine-tuned methods, non-fine-tuned LLM-type methods, and our \texttt{TMPT}.}
\label{results-in-target}
\end{table*}

\section{Experimental Setup}
To advance and facilitate research in the field of multi-modal stance detection, based on our new datasets, we conduct experiments on \textbf{In-target Multi-modal Stance Detection}, training and testing on the same target, and \textbf{Zero-shot Multi-modal Stance Detection}, performing stance detection on unseen targets based on the known targets. 

\subsection{Data Partition}
\label{subsec:data_partition}
For in-target multi-modal stance detection, each dataset is divided into training, development, and testing sets with a ratio of 7:1:2\footnote{To address the possibility of one text corresponding to multiple images, we use the Twitter ID to split datasets, thus avoiding the issue of data leakage.}.
For zero-shot multi-modal stance detection, the training and development set is built on known target(s), and the testing set is built on unknown target(s). Since there is only one target in the \textsc{Mccq} dataset, we only use it for in-target multi-modal stance detection. In the \textsc{Mwtwt} dataset, following~\cite{conforti-etal-2020-will}, we select four targets (CVS\_AET, CI\_ESRX, ANTM\_CI, and AET\_HUM) to perform zero-shot scenario since DIS\_FOXA is not in the same domain as them.
The statistics of each dataset are shown in Table~\ref{dataset-statistic}.

\subsection{Comparison Models}
{\bf Pure textual modality baselines}: 
1) BERT~\cite{devlin-etal-2019-bert}, the uncased BERT-base;
2) RoBERTa~\cite{DBLP:journals/corr/abs-1907-11692}, the RoBERTa-base; 
3) KEBERT~\cite{DBLP:conf/lrec/KawintiranonS22}, a BERTweet-base model with specific knowledge of Twitter political posts. 
4) LLaMA2~\cite{DBLP:journals/corr/abs-2307-09288}, the LLaMA2-70b-chat;
5) GPT4~\footnote{\url{https://openai.com/research/gpt-4}}.
{\bf Pure visual modality baselines}: 
1) ResNet~\cite{DBLP:conf/cvpr/HeZRS16}, the ResNet-50 v1.5; 2) ViT~\cite{dosovitskiy2021an}, the vit-base-patch16-224; 3) Swin Transformer (SwinT)~\cite{liu2021swin}, the swinv2-base-patch4-window12-192-22k.
{\bf Multi-modal baselines}: 
1) ViLT~\cite{kim2021vilt}, the vilt-b32-mlm;
2) CLIP~\cite{radford2021learning}, the clip-vit-base-patch32;
3) BERT+ViT, utilizing BERT as the textual encoder and ViT as the visual encoder, and concatenating the \texttt{[CLS]} vectors of textual and visual modalities for stance detection.
4) Qwen-VL~\cite{DBLP:journals/corr/abs-2308-12966}, the Qwen-VL-Chat-7b.
5) GPT4-Vision~\footnote{\url{https://openai.com/research/gpt-4v-system-card}}.

\subsection{Experimental Settings}
To leverage the powerful capabilities of large language models, following~\citet{DBLP:conf/emnlp/GattoSP23}, we propose a variant of our TMPT model, named \texttt{TMPT}+\texttt{CoT}. By utilizing GPT4-Vision to generate a chain of thought from the text and image of the sample, we concatenate this chain of thought with the text to serve as the textual modality input for \texttt{TMPT}+\texttt{CoT}. The images are used as the visual modality input. This approach is employed for both the training and testing phases.
We utilize the pre-trained uncased BERT-base~\cite{devlin-etal-2019-bert} to embed each word as a 768-dimensional embedding and employ the pre-trained ViT-base~\cite{dosovitskiy2021an} to embed each image patch as a 768-dimensional embedding, \textit{i.e.}, $d^T=d^V=768$. The resolution of the visual patch is set to $(16,16)$. The length of visual prompt tokens is set to $c=7$.
The dimensionality of hidden vectors is set to $d_h=768$. 
We use {\em Macro F1-score} to measure the model performance. The experimental results of our models are averaged over 5 runs to ensure the final reported results are statistically stable. For detailed settings of the experiments, please refer to Appendix~\ref{prompt-tuning-analysis}.

\begin{table*}[!t]
\small
\centering
\setlength{\tabcolsep}{2.5pt}
\renewcommand{\arraystretch}{1.1}
\begin{tabular}{llccccccccccccc}
\hline
\multirow{2}*{\textsc{Modality}}         & \multirow{2}*{\textsc{Method}}    & \multicolumn{2}{c}{MTSE}        &  & \multicolumn{4}{c}{MWTWT}                                         &  & \multicolumn{2}{c}{MRUC}        &  & \multicolumn{2}{c}{MTWQ}        \\
\cline{3-4} \cline{6-9} \cline{11-12} \cline{14-15}
                             &                        & DT             & JB             &  & CA             & CE             & AC             & AH             &  & RUS            & UKR            &  & MOC            & TOC            \\
\hline
\multirow{4}{*}{Textual}     & BERT                   & 32.52          & 29.97          &  & 63.55          & 61.30          & 59.18          & 52.89          &  & 22.01          & 15.45          &  & 28.04          & 9.57          \\
                             & RoBERTa                & 26.60          & 32.21          &  & 59.22          & 59.22          & 64.86          & 57.46          &  & 27.10          & 19.98          &  & 30.62          & 15.84          \\
                             & KEBERT                 & 26.17          & 31.81          &  & 59.70          & 62.56          & 63.92          & 55.53          &  & 24.68          & 28.18          &  & 29.17          & 19.80          \\
                             \cdashline{2-15}[2pt/3pt]
                             & LLaMA2                 & 53.57          & 53.92          &  & 32.47          & 38.37          & 48.08          & 46.13          &  & 31.86          & 36.34          &  & 51.46          & 44.10          \\
                             & GPT4                   & 70.78          & 68.83          &  & 57.19          & 60.56          & 65.63          & \textbf{69.01} &  & 40.22          & 49.18          &  & 62.10          & 52.12          \\
\hline
\multirow{3}{*}{Visual}      & ResNet                 & 25.52          & 29.70          &  & 23.01          & 24.11          & 25.21          & 25.27          &  & 23.88          & 25.57          &  & 27.59          & 24.88          \\
                             & ViT                    & 28.63          & 29.70          &  & 24.59          & 28.18          & 34.06          & 33.40          &  & 27.26          & 28.51          &  & 29.37          & 23.69          \\
                             & SwinT       & 28.54          & 30.85          &  & 28.53          & 28.50          & 35.87          & 34.33          &  & 25.44          & 24.54          &  & 27.90          & 19.69          \\
\hline
\multirow{7}{*}{Multi-modal} & BERT+ViT               & 26.70          & 31.57          &  & 59.21          & 59.30          & 65.04          & 59.28          &  & 23.33          & 15.21          &  & 24.76          & 11.70          \\
                             & ViLT                   & 28.08          & 29.74          &  & 38.33          & 46.00          & 55.01          & 48.55          &  & 21.56          & 23.96          &  & 23.54          & 19.18          \\
                             & CLIP                   & 28.21          & 28.99          &  & 61.08          & 55.67          & 63.80          & 60.06          &  & 25.62          & 27.40          &  & 27.21          & 15.69          \\
                             \cdashline{2-15}[2pt/3pt]
                             & Qwen-VL                & 47.62          & 46.14          &  & 38.57          & 43.36          & 47.82          & 41.01          &  & 36.95          & 41.39          &  & 44.32          & 44.08          \\
                             & GPT4-Vision                 & \textbf{72.68} & \textbf{71.28} &  & 42.23          & 45.92          & 54.59          & 53.19          &  & 42.09          & 47.00          &  & \textbf{65.00}          & \textbf{52.36} \\
                             \cdashline{2-15}[2pt/3pt]
                              & \cellcolor[gray]{0.92}\texttt{TMPT}       & \cellcolor[gray]{0.92}31.69 & \cellcolor[gray]{0.92}32.65 & \cellcolor[gray]{0.92} & \cellcolor[gray]{0.92}66.36          & \cellcolor[gray]{0.92}\textbf{66.39}$^\star$          & \cellcolor[gray]{0.92}\textbf{66.32} & \cellcolor[gray]{0.92}61.56 & \cellcolor[gray]{0.92} & \cellcolor[gray]{0.92}23.87          & \cellcolor[gray]{0.92}24.71          & \cellcolor[gray]{0.92} & \cellcolor[gray]{0.92}32.18 & \cellcolor[gray]{0.92}26.48 \\
& \cellcolor[gray]{0.92}\texttt{TMPT}+\texttt{CoT} & \cellcolor[gray]{0.92}54.30 & \cellcolor[gray]{0.92}58.46 & \cellcolor[gray]{0.92} & \cellcolor[gray]{0.92}\textbf{67.28}$^\star$ & \cellcolor[gray]{0.92}63.73 & \cellcolor[gray]{0.92}64.87          & \cellcolor[gray]{0.92}54.26 & \cellcolor[gray]{0.92} & \cellcolor[gray]{0.92}\textbf{48.99}$^\star$ & \cellcolor[gray]{0.92}\textbf{51.75}$^\star$ & \cellcolor[gray]{0.92} & \cellcolor[gray]{0.92}45.32 & \cellcolor[gray]{0.92}43.70 \\
\hline
\end{tabular}
\caption{Experimental results of zero-shot multi-modal stance detection. The dark background results are for our \texttt{TMPT}. Best scores of each group are in bold. Results with $\star$ denote the significance tests of our \texttt{TMPT} over the baseline models at $p\rm{-}value<0.05$. The dashed line represents a separation between fine-tuned methods, non-fine-tuned LLM-type methods, and our \texttt{TMPT}.}
\label{results-zero-shot}
\end{table*}

\section{Experimental Results}

\subsection{In-target Multi-modal Stance Detection}
The results of in-target multi-modal stance detection are shown in Table~\ref{results-in-target}. It can be seen that our proposed \texttt{TMPT} outperforms fine-tuning based baselines on most of datasets, denoting that exploiting targeted prompt tuning can preferably leverage the target-specific multi-modal stance information, thus improving stance detection performance. Further, \texttt{TMPT}+\texttt{CoT} performs overall better than \texttt{TMPT}, which demonstrates that leveraging the knowledge from large language models can improve the comprehension of textual modality, thereby achieving better performance.
For uni-modal baselines, the performance of visual modality methods is unsatisfactory, while the performance of textual modality methods is much better. This indicates that the stance expression primarily resides in the textual modality.
Further, CLIP, which considers both textual and visual modalities performs overall better than models which only consider textual modality, which proves the importance of visual modality in multi-modal stance detection. While, in datasets like \textsc{Mwtwt} and \textsc{Mccq}, GPT4-Vision underperforms GPT-4. Our analysis of images within these datasets revealed that a noteworthy proportion of images had comparatively high complexity levels. 
This also implies effective use of visual information is key to improving multi-modal stance detection performance.
In addition, KEBERT, which integrates specific knowledge of Twitter political posts into BERTweet, achieves promising performance compared to other textual models, which indicates that exploring external target-related knowledge might improve the performance of multi-modal stance detection.

\subsection{Zero-shot Multi-modal Stance Detection}
The results of zero-shot multi-modal stance detection are reported in Table~\ref{results-zero-shot}. It can be seen that the large language models achieve superior performance due to the need for detecting stances on unseen targets. This may be attributed to the powerful zero-shot learning ability of large language models.
For our \texttt{TMPT}, which does not use chain-of-thoughts from large language models, still achieves better performance than large language models in some targets, while also outperforming most of non-large language model baselines. This shows the promise of our \texttt{TMPT} in zero-shot stance detection. Further, \texttt{TMPT}+\texttt{CoT} performs overall better than \texttt{TMPT}, which indicates that obtaining powerful text and visual comprehension abilities from large models may be key to improving the performance of detecting stance for unseen targets.

\begin{table}[!t]
\small
\centering
\setlength{\tabcolsep}{2.5pt}
\renewcommand{\arraystretch}{1.1}
\begin{tabular}{l|ccccc}
\hline
 {\textsc{Method}} & MTSE & MCCQ & MWTWT & MRUC & MTWQ  \\
 \hline
\cellcolor[gray]{0.92}\texttt{TMPT} & \cellcolor[gray]{0.92}{\bf 60.84} & \cellcolor[gray]{0.92}{\bf 67.67} & \cellcolor[gray]{0.92}{\bf 77.59} & \cellcolor[gray]{0.92}{\bf 75.76} & \cellcolor[gray]{0.92}{\bf 67.59}\\
\cdashline{1-4}[2pt/3pt]
w/o $\mathcal{P}^T$ & 54.93 & 62.76 & 71.42 & 70.53 & 61.77 \\
w/o $\mathcal{P}^V$ & 58.14 & 65.71 & 73.93 & 70.85 & 63.69 \\
 \hline
\end{tabular}
\caption{Experimental results of ablation study. The reported results are the Macro F1-score across all targets in a dataset on in-target multi-modal stance detection.}
\label{results-ablation}
\end{table}

\subsection{Ablation Study}
\label{ablation:study}
To analyze the impact of the targeted prompt tuning in our proposed \texttt{TMPT}, we conduct an ablation study and report the results in Table~\ref{results-ablation}. Note that the removal of textual prompting (w/o $\mathcal{P}^T$) sharply degrades the performance, which verifies the significance of textual prompting in learning textual targeted stance features for multi-modal stance detection. In addition, the removal of visual prompt tuning (w/o $\mathcal{P}^V$) leads to considerable performance degradation, which indicates that utilizing visual prompt tuning can make better learning of visual targeted stance information and thus improves the performance of multi-modal stance detection.

\begin{figure}[!t]
\centering
\begin{subfigure}[b]{0.5\textwidth}
    \centering
    \includegraphics[width=0.8\textwidth]{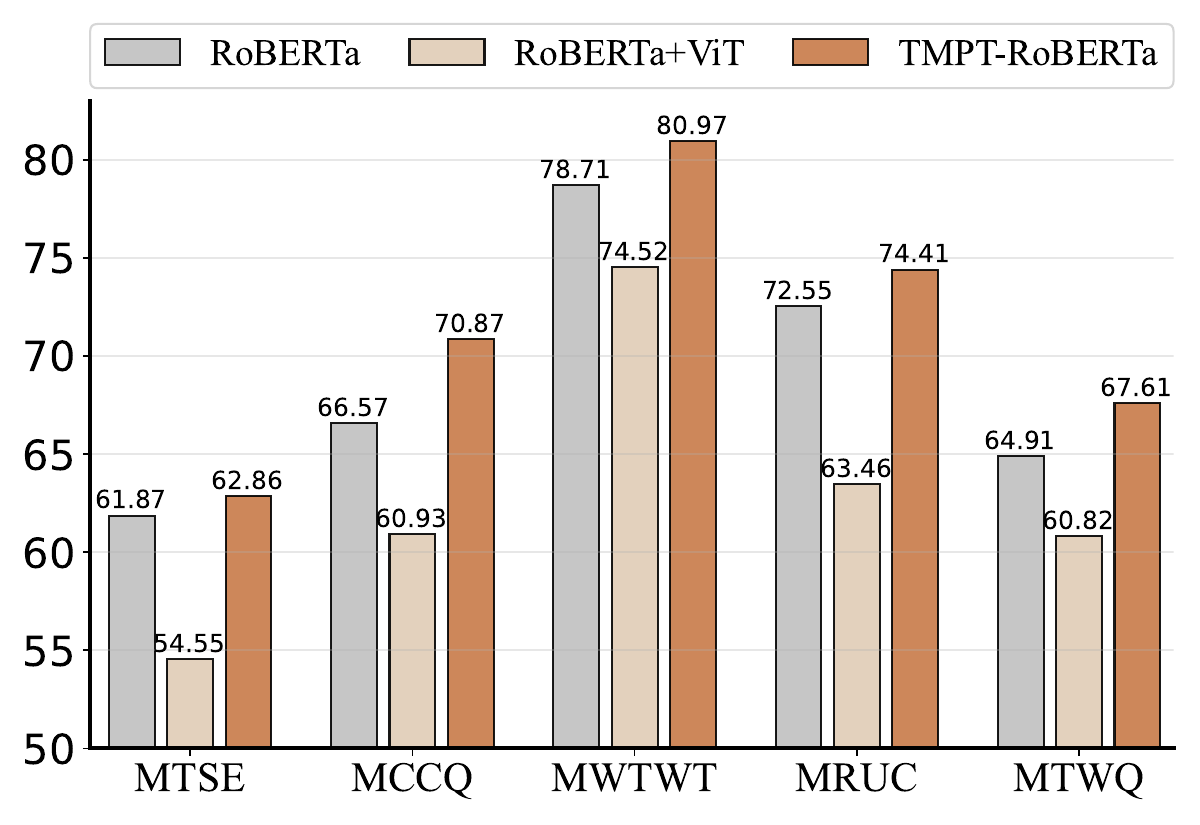}
\end{subfigure}
\begin{subfigure}[b]{0.5\textwidth}
    \centering
    \includegraphics[width=0.8\textwidth]{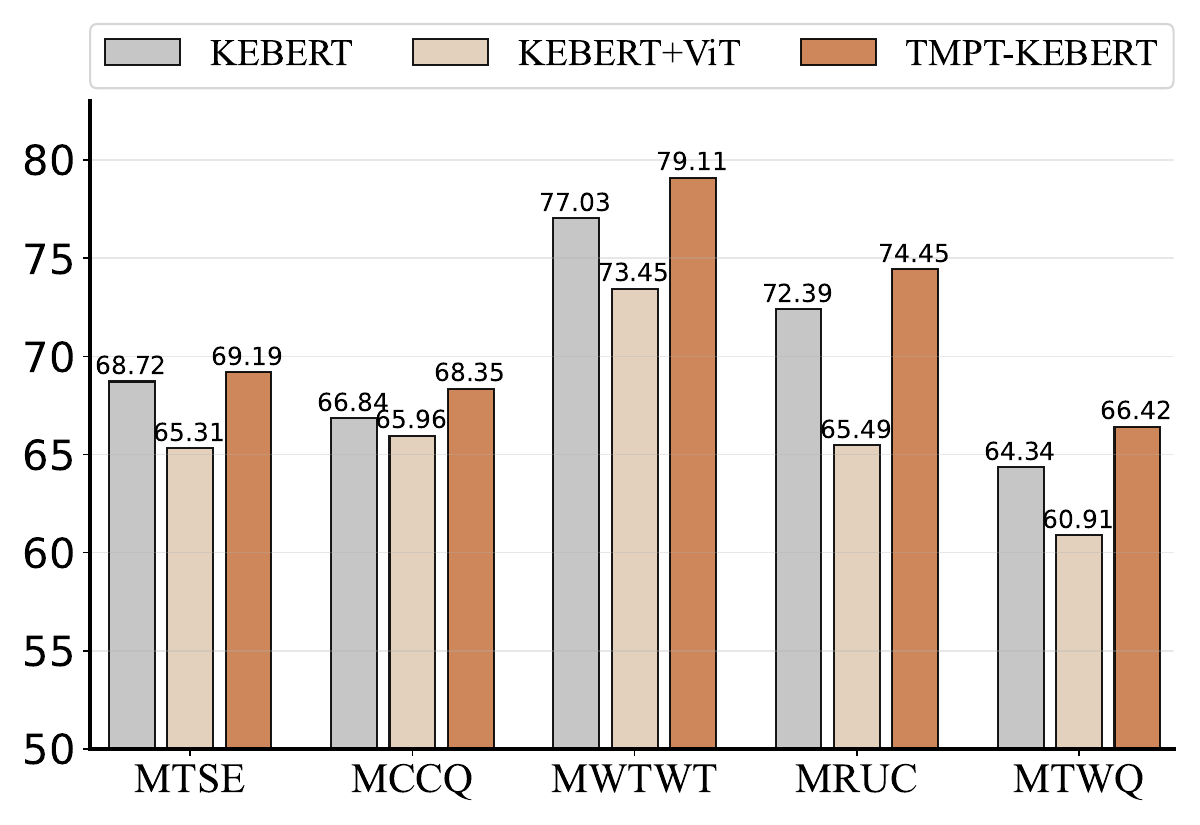}
\end{subfigure}
\caption{Performance of using different pre-trained language models: RoBERTa~\cite{DBLP:conf/emnlp/NguyenVN20} (top) and KEBERT~\cite{kawintiranon-singh-2021-knowledge} (bottom). The reported results are the Macro F1-score across all targets in a dataset on in-target multi-modal stance detection.}
\label{fig-pre}
\end{figure}

\begin{figure}[!t]
\centering  
\includegraphics[width=0.95\linewidth]{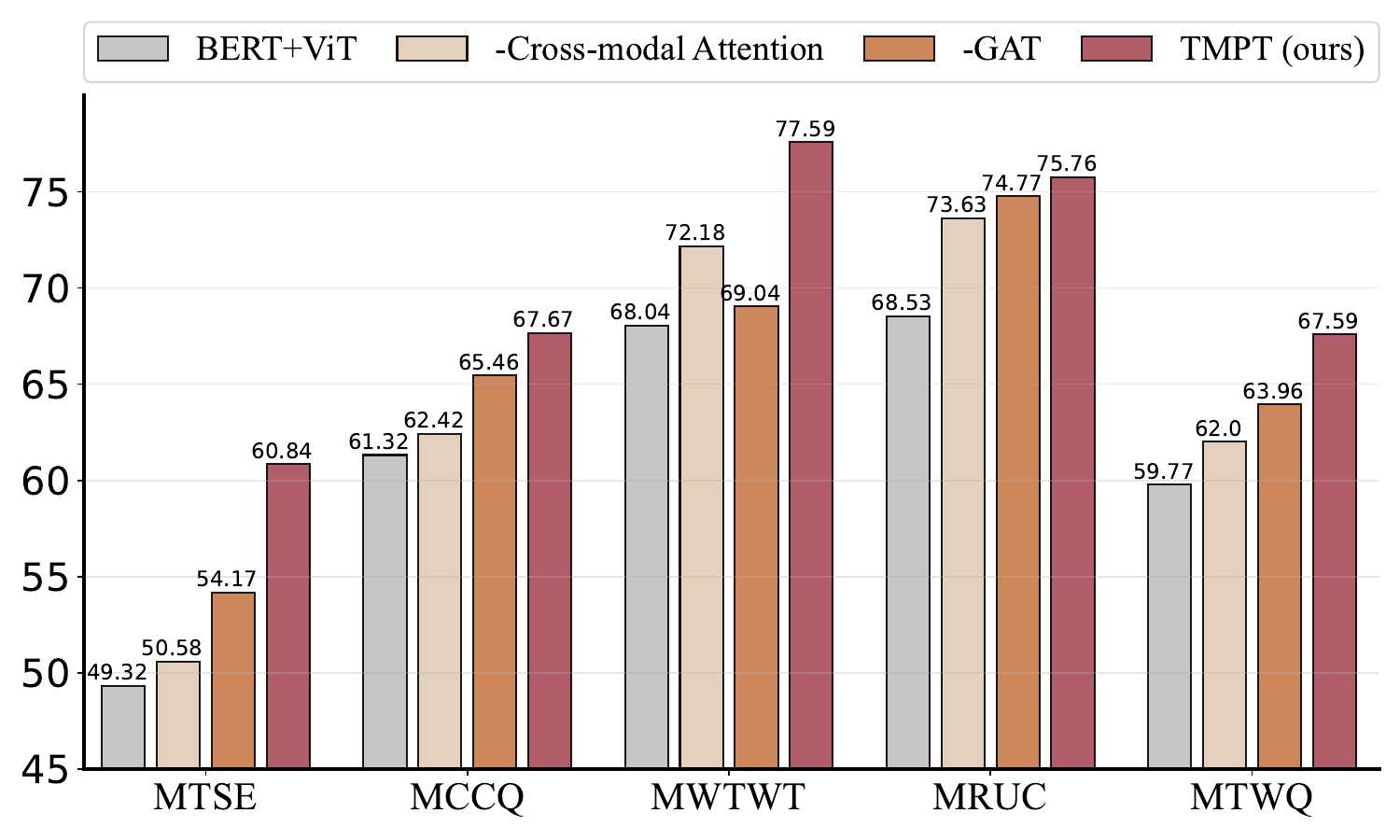}
\caption{Performance of using different multi-modal fusion methods. The reported results are the Macro F1-score across all targets in a dataset on in-target multi-modal stance detection.}
\label{fig-fusion}
\end{figure}

\subsection{Generalization of Targeted Multi-modal Prompt Tuning}
\paragraph{Pre-trained Models}Previous experiments have demonstrated that the stance expression primarily resides in the textual modality. Therefore, to investigate the generalization of our \texttt{TMPT} when used with different pre-trained language models, we conduct experiments with two variants of our \texttt{TMPT} by using two other promising PLMs: RoBERTa~\cite{liu2019roberta} and KEBERT~\cite{kawintiranon-singh-2021-knowledge}. The results are shown in Figure~\ref{fig-pre}. Note that our Targeted Multi-modal Prompting can directly work with the two PLMs and achieve better performance, showing the compatibility of our method with various pre-trained models.

\paragraph{Multi-modal Fusion}
We choose cross-modal attention~\cite{wei2020multi} and GAT~\cite{veličković2018graph} to analyze the performance of our \texttt{TMPT} with different multi-modal fusion methods. The results are shown in Figure~\ref{fig-fusion}. It can be seen that no matter which fusion method we use, the experimental performance is better than BERT+ViT. This indicates that our \texttt{TMPT} can directly work with various multi-modal fusion methods and lead to improved multi-modal stance detection performance.

\subsection{Error Analysis}
From the results in Table~\ref{results-zero-shot}, we can see that our \texttt{TMPT} performs well in the \textsc{Mwtwt} dataset, but still insufficient in other datasets on zero-shot stance detection. One possible reason is that targets in \textsc{Mwtwt} are all about expressing views on corporate mergers, so unknown targets can easily find commonalities in the dataset. However, for other datasets, the topics are diverse, which poses a challenge to mining targeted information. Therefore, how to better learn the correlation information between targets from data on diverse topics is a potential direction to improve the performance of the zero-shot scenario.

Further, after analyzing examples of misclassification, we found that among the incorrect samples, approximately 70\% (based on calculations on randomly sampled 300 incorrect samples from five datasets) of the images contained stance-related content. This indicates that for our multi-modal stance detection task, it is important for further exploration in extracting and utilizing features from the visual modality.

\begin{table}[t]
\small
\centering
\setlength{\tabcolsep}{2.5pt}
\renewcommand{\arraystretch}{1.1}
\begin{tabular}{l|ccccc}
\hline
      & Person & Events & Words & Memes & Mixed \\
\hline
ERR-Img(\%)  & 7.8   & 13.4    & 37.7  & 26.6  & 14.5  \\
ERR-Img-prop  & 0.28   & 0.60   & 1.60  & 1.95  & 1.16  \\
\hline
\end{tabular}
\caption{The results of error analysis on image types. ERR-Img means the proportion of different image types among the incorrect samples with stance-related content. ERR-Img-prop means ERR-Img divided by the proportion of its image types in the original dataset.}
\label{tab:image_error}
\end{table}

Building on the previous step, we conduct error analysis on different image types. The results are illustrated in Table~\ref{tab:image_error}. A larger ERR-Img-prop indicates a weaker ability of the model to handle samples with images of that category, indicating that for images containing words, memes, and mixed features, more effective methods of feature extraction and understanding remain to be proposed.

\subsection{Visualization}
\label{visual}

\begin{figure}[!t]
\centering  
\includegraphics[width=0.9\linewidth]{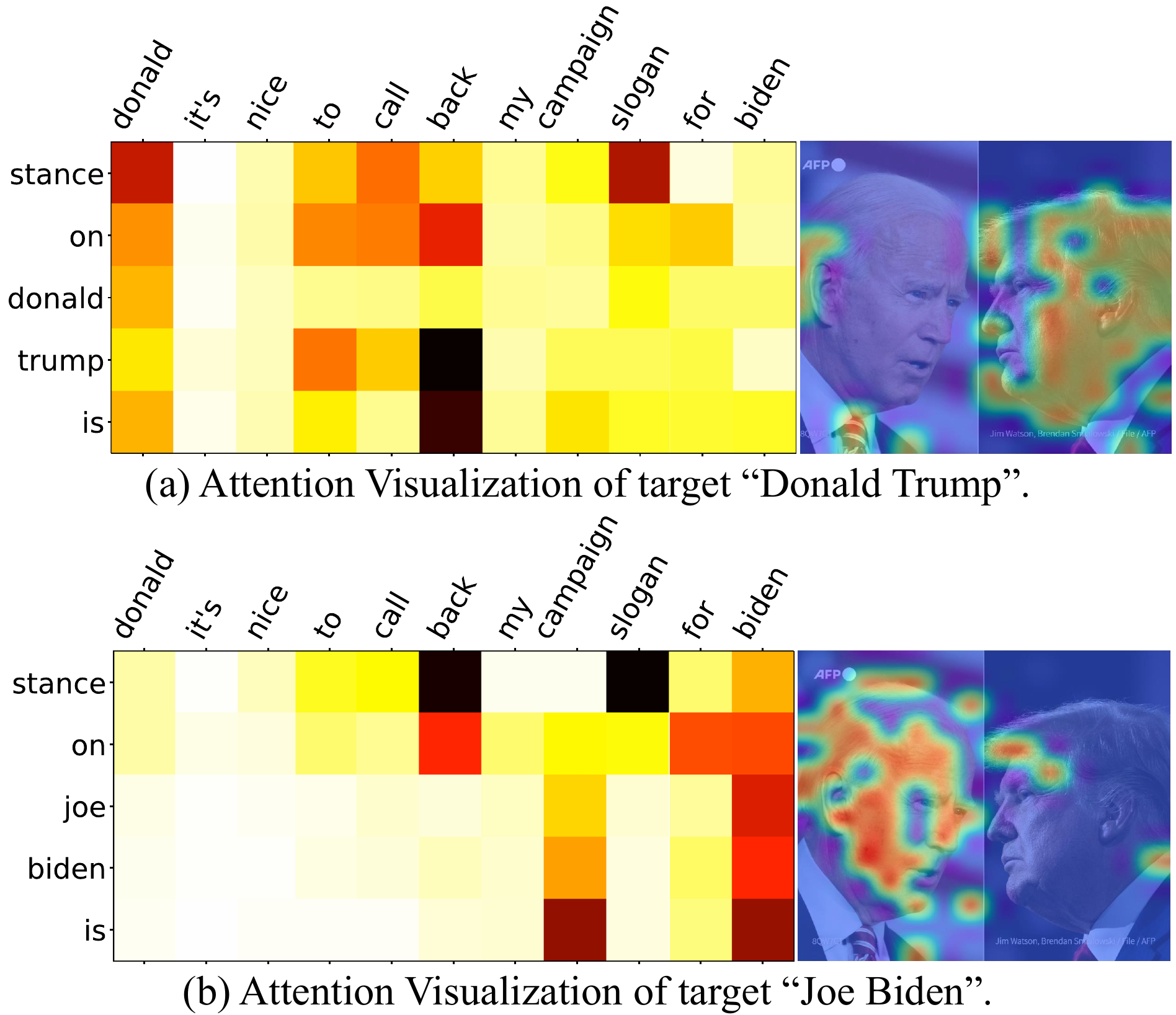}
\caption{Visualization of a typical example.}
\label{fig-visualization}
\end{figure}

To qualitatively investigate how our \texttt{TMPT} improves the performance of multi-modal stance detection, we visualize the attention values calculated by the targeted prompts and the vectors of the final layer of two encoders. The results are shown in Figure~\ref{fig-visualization}. It can be seen that the crucial textual tokens and the key visual patches regarding different targets are highly attended to and discriminated by our \texttt{TMPT}. This illustrates that the proposed \texttt{TMPT} can learn the important stance features for the specific target with the help of targeted prompt tuning, thus improving the learning ability of multi-modal stance detection.

\section{Conclusion}
In this paper, we present \textsc{Mtse}, \textsc{Mccq}, \textsc{Mwtwt}, \textsc{Mruc} and \textsc{Mtwq}, five new datasets for multi-modal stance detection. Based on the created datasets, we present in-target multi-modal stance detection and zero-shot multi-modal stance detection, aiming to advance and facilitate research in the field of multi-modal stance detection. In addition, we propose a simple yet effective targeted multi-modal prompting framework (\texttt{TMPT}) to deal with multi-modal stance detection, where the target information is explored to prompt the pre-trained models in learning multi-modal stance features.
Extensive experiments on the new datasets show that our proposed \texttt{TMPT} achieves overall better performance than state-of-the-art baseline methods.

\section*{Acknowledgements}
This work was partially supported by the National Natural Science Foundation of China (62176076),  Natural Science Foundation of GuangDong 2023A1515012922, the Shenzhen Foundational Research Funding (JCYJ20220818102415032, JCYJ20210324115614039), the Major Key Project of PCL2021A06, Guangdong Provincial Key Labo-ratory of Novel Security Intelligence Technologies 2022B1212010005.

\section*{Limitations}
Our method needs to devise a specific prompt for the given target, which needs to take time and artificial effort to analyze the target information in the real-world scenario for designing and selecting appropriate prompts. Furthermore, our proposed method does not integrate external target-specific knowledge to improve the learning of multi-modal stance information, such as the background knowledge of targets. Integrating external knowledge related to the target can improve the performance of stance detection. In addition, the current version of the data does not consider audio modality and video information, which is also an issue we need to explore in the future.

\section*{Ethics Statement}
This work presents \textsc{Mtse}, \textsc{Mccq}, \textsc{Mwtwt}, \textsc{Mruc} and \textsc{Mtwq}, five new open-source datasets for the research community to study multi-modal stance detection. The \textsc{Mtse}, \textsc{Mccq}, and \textsc{Mwtwt} are the extension of Twitter Stance Election 2020~\cite{kawintiranon-singh-2021-knowledge}, COVID-CQ \cite{mutlu2020stance}, and Will-They-Won’t-They \cite{conforti-etal-2020-will}, which are three open-source textual stance detection datasets for academic research.
We only collect the tweet text and image content needed for our research from Twitter following the privacy agreement of Twitter for academic usage, so there is no privacy issue.
To annotate extended data, we recruited 8 experienced researchers who work on natural language processing or multi-modal learning. The detailed collect and annotate process has been illustrated in Section~\ref{data_details}.
Each researcher is paid \$6.5 per hour (above the average local payment of similar jobs). The entire annotation process lasted 5 months, and the average annotation time of the eight researchers was 430 hours.
During the annotation process, samples that contain personally identifiable information will be discarded, and only the tweet IDs and human-annotated stance labels will be shared. Thus, our data set complies with Twitter's information privacy policy. The annotators have no affiliation with any of the companies that are used as targets in the dataset, so there is no potential bias due to conflict of interest. We used the ChatGPT service from OpenAI for our writing. We followed their term and policies.
Some examples in our paper may include a stance or tendency. It should be clarified that they are randomly sampled from the dataset for better studying the dataset and task, and do not represent any personal viewpoints.

\bibliography{acl_latex}

\begin{thebibliography}{50}
\expandafter\ifx\csname natexlab\endcsname\relax\def\natexlab#1{#1}\fi

\bibitem[{Allaway and McKeown(2020)}]{allaway2020zero}
Emily Allaway and Kathleen McKeown. 2020.
\newblock \href {https://doi.org/10.18653/v1/2020.emnlp-main.717} {{Z}ero-{S}hot {S}tance {D}etection: {A} {D}ataset and {M}odel using {G}eneralized {T}opic {R}epresentations}.
\newblock In \emph{Proceedings of the 2020 Conference on Empirical Methods in Natural Language Processing (EMNLP)}, pages 8913--8931, Online. Association for Computational Linguistics.

\bibitem[{Allaway et~al.(2021)Allaway, Srikanth, and McKeown}]{allaway-etal-2021-adversarial}
Emily Allaway, Malavika Srikanth, and Kathleen McKeown. 2021.
\newblock \href {https://doi.org/10.18653/v1/2021.naacl-main.379} {Adversarial learning for zero-shot stance detection on social media}.
\newblock In \emph{Proceedings of the 2021 Conference of the North American Chapter of the Association for Computational Linguistics: Human Language Technologies}, pages 4756--4767, Online. Association for Computational Linguistics.

\bibitem[{Augenstein et~al.(2016)Augenstein, Rockt{\"a}schel, Vlachos, and Bontcheva}]{augenstein-etal-2016-stance}
Isabelle Augenstein, Tim Rockt{\"a}schel, Andreas Vlachos, and Kalina Bontcheva. 2016.
\newblock \href {https://doi.org/10.18653/v1/D16-1084} {Stance detection with bidirectional conditional encoding}.
\newblock In \emph{Proceedings of the 2016 Conference on Empirical Methods in Natural Language Processing}, pages 876--885, Austin, Texas. Association for Computational Linguistics.

\bibitem[{Bai et~al.(2023)Bai, Bai, Yang, Wang, Tan, Wang, Lin, Zhou, and Zhou}]{DBLP:journals/corr/abs-2308-12966}
Jinze Bai, Shuai Bai, Shusheng Yang, Shijie Wang, Sinan Tan, Peng Wang, Junyang Lin, Chang Zhou, and Jingren Zhou. 2023.
\newblock \href {https://doi.org/10.48550/ARXIV.2308.12966} {Qwen-vl: {A} frontier large vision-language model with versatile abilities}.
\newblock \emph{CoRR}, abs/2308.12966.

\bibitem[{Chen et~al.(2021)Chen, Ye, and Cui}]{chen2021integrating}
Pengyuan Chen, Kai Ye, and Xiaohui Cui. 2021.
\newblock Integrating n-gram features into pre-trained model: a novel ensemble model for multi-target stance detection.
\newblock In \emph{International Conference on Artificial Neural Networks}, pages 269--279. Springer.

\bibitem[{Cohen(1960)}]{cohen1960coefficient}
Jacob Cohen. 1960.
\newblock A coefficient of agreement for nominal scales.
\newblock \emph{Educational and psychological measurement}, 20(1):37--46.

\bibitem[{Conforti et~al.(2020)Conforti, Berndt, Pilehvar, Giannitsarou, Toxvaerd, and Collier}]{conforti-etal-2020-will}
Costanza Conforti, Jakob Berndt, Mohammad~Taher Pilehvar, Chryssi Giannitsarou, Flavio Toxvaerd, and Nigel Collier. 2020.
\newblock \href {https://doi.org/10.18653/v1/2020.acl-main.157} {Will-they-won{'}t-they: A very large dataset for stance detection on {T}witter}.
\newblock In \emph{Proceedings of the 58th Annual Meeting of the Association for Computational Linguistics}, pages 1715--1724, Online. Association for Computational Linguistics.

\bibitem[{Devlin et~al.(2019)Devlin, Chang, Lee, and Toutanova}]{devlin-etal-2019-bert}
Jacob Devlin, Ming-Wei Chang, Kenton Lee, and Kristina Toutanova. 2019.
\newblock \href {https://doi.org/10.18653/v1/N19-1423} {{BERT}: Pre-training of deep bidirectional transformers for language understanding}.
\newblock In \emph{Proceedings of the 2019 Conference of the North {A}merican Chapter of the Association for Computational Linguistics: Human Language Technologies, Volume 1 (Long and Short Papers)}, pages 4171--4186, Minneapolis, Minnesota. Association for Computational Linguistics.

\bibitem[{Dosovitskiy et~al.(2021)Dosovitskiy, Beyer, Kolesnikov, Weissenborn, Zhai, Unterthiner, Dehghani, Minderer, Heigold, Gelly, Uszkoreit, and Houlsby}]{dosovitskiy2021an}
Alexey Dosovitskiy, Lucas Beyer, Alexander Kolesnikov, Dirk Weissenborn, Xiaohua Zhai, Thomas Unterthiner, Mostafa Dehghani, Matthias Minderer, Georg Heigold, Sylvain Gelly, Jakob Uszkoreit, and Neil Houlsby. 2021.
\newblock \href {https://openreview.net/forum?id=YicbFdNTTy} {An image is worth 16x16 words: Transformers for image recognition at scale}.
\newblock In \emph{9th International Conference on Learning Representations, {ICLR} 2021, Virtual Event, Austria, May 3-7, 2021}. OpenReview.net.

\bibitem[{Ebrahimi et~al.(2016)Ebrahimi, Dou, and Lowd}]{ebrahimi-etal-2016-weakly}
Javid Ebrahimi, Dejing Dou, and Daniel Lowd. 2016.
\newblock \href {https://doi.org/10.18653/v1/D16-1105} {Weakly supervised tweet stance classification by relational bootstrapping}.
\newblock In \emph{Proceedings of the 2016 Conference on Empirical Methods in Natural Language Processing}, pages 1012--1017, Austin, Texas. Association for Computational Linguistics.

\bibitem[{Gatto et~al.(2023)Gatto, Sharif, and Preum}]{DBLP:conf/emnlp/GattoSP23}
Joseph Gatto, Omar Sharif, and Sarah Preum. 2023.
\newblock \href {https://aclanthology.org/2023.findings-emnlp.273} {Chain-of-thought embeddings for stance detection on social media}.
\newblock In \emph{Findings of the Association for Computational Linguistics: {EMNLP} 2023, Singapore, December 6-10, 2023}, pages 4154--4161. Association for Computational Linguistics.

\bibitem[{Hasan and Ng(2013)}]{hasan-ng-2013-stance}
Kazi~Saidul Hasan and Vincent Ng. 2013.
\newblock \href {https://aclanthology.org/I13-1191} {Stance classification of ideological debates: Data, models, features, and constraints}.
\newblock In \emph{Proceedings of the Sixth International Joint Conference on Natural Language Processing}, pages 1348--1356, Nagoya, Japan. Asian Federation of Natural Language Processing.

\bibitem[{Hasan and Ng(2014)}]{hasan-ng-2014-taking}
Kazi~Saidul Hasan and Vincent Ng. 2014.
\newblock \href {https://doi.org/10.3115/v1/D14-1083} {Why are you taking this stance? identifying and classifying reasons in ideological debates}.
\newblock In \emph{Proceedings of the 2014 Conference on Empirical Methods in Natural Language Processing ({EMNLP})}, pages 751--762, Doha, Qatar. Association for Computational Linguistics.

\bibitem[{He et~al.(2016)He, Zhang, Ren, and Sun}]{DBLP:conf/cvpr/HeZRS16}
Kaiming He, Xiangyu Zhang, Shaoqing Ren, and Jian Sun. 2016.
\newblock \href {https://doi.org/10.1109/CVPR.2016.90} {Deep residual learning for image recognition}.
\newblock In \emph{2016 {IEEE} Conference on Computer Vision and Pattern Recognition, {CVPR} 2016, Las Vegas, NV, USA, June 27-30, 2016}, pages 770--778. {IEEE} Computer Society.

\bibitem[{Jia et~al.(2022)Jia, Tang, Chen, Cardie, Belongie, Hariharan, and Lim}]{jia2022vpt}
Menglin Jia, Luming Tang, Bor-Chun Chen, Claire Cardie, Serge Belongie, Bharath Hariharan, and Ser-Nam Lim. 2022.
\newblock Visual prompt tuning.
\newblock In \emph{European Conference on Computer Vision (ECCV)}.

\bibitem[{Jiang et~al.(2020)Jiang, Xu, Araki, and Neubig}]{Jiang-etal-2020-How}
Zhengbao Jiang, Frank~F. Xu, Jun Araki, and Graham Neubig. 2020.
\newblock \href {https://doi.org/10.1162/tacl_a_00324} {How can we know what language models know?}
\newblock \emph{Transactions of the Association for Computational Linguistics}, 8:423--438.

\bibitem[{Ju et~al.(2022)Ju, Han, Zheng, Zhang, and Xie}]{Ju-etal-2022-Prompting}
Chen Ju, Tengda Han, Kunhao Zheng, Ya~Zhang, and Weidi Xie. 2022.
\newblock \href {https://doi.org/10.1007/978-3-031-19833-5\_7} {Prompting visual-language models for efficient video understanding}.
\newblock In \emph{Computer Vision - {ECCV} 2022 - 17th European Conference, Tel Aviv, Israel, October 23-27, 2022, Proceedings, Part {XXXV}}, volume 13695 of \emph{Lecture Notes in Computer Science}, pages 105--124. Springer.

\bibitem[{Kawintiranon and Singh(2021)}]{kawintiranon-singh-2021-knowledge}
Kornraphop Kawintiranon and Lisa Singh. 2021.
\newblock \href {https://doi.org/10.18653/v1/2021.naacl-main.376} {Knowledge enhanced masked language model for stance detection}.
\newblock In \emph{Proceedings of the 2021 Conference of the North American Chapter of the Association for Computational Linguistics: Human Language Technologies}, pages 4725--4735, Online. Association for Computational Linguistics.

\bibitem[{Kawintiranon and Singh(2022)}]{DBLP:conf/lrec/KawintiranonS22}
Kornraphop Kawintiranon and Lisa Singh. 2022.
\newblock \href {https://aclanthology.org/2022.lrec-1.801} {Polibertweet: {A} pre-trained language model for analyzing political content on twitter}.
\newblock In \emph{Proceedings of the Thirteenth Language Resources and Evaluation Conference, {LREC} 2022, Marseille, France, 20-25 June 2022}, pages 7360--7367. European Language Resources Association.

\bibitem[{Kim et~al.(2021)Kim, Son, and Kim}]{kim2021vilt}
Wonjae Kim, Bokyung Son, and Ildoo Kim. 2021.
\newblock \href {http://proceedings.mlr.press/v139/kim21k.html} {Vilt: Vision-and-language transformer without convolution or region supervision}.
\newblock In \emph{Proceedings of the 38th International Conference on Machine Learning, {ICML} 2021, 18-24 July 2021, Virtual Event}, volume 139 of \emph{Proceedings of Machine Learning Research}, pages 5583--5594. {PMLR}.

\bibitem[{Lester et~al.(2021)Lester, Al-Rfou, and Constant}]{Lester-etal-2021-The-Power}
Brian Lester, Rami Al-Rfou, and Noah Constant. 2021.
\newblock \href {https://doi.org/10.18653/v1/2021.emnlp-main.243} {The power of scale for parameter-efficient prompt tuning}.
\newblock In \emph{Proceedings of the 2021 Conference on Empirical Methods in Natural Language Processing}, pages 3045--3059, Online and Punta Cana, Dominican Republic. Association for Computational Linguistics.

\bibitem[{Li et~al.(2023)Li, Liang, Zhao, Zhang, Yang, and Xu}]{DBLP:conf/emnlp/LiLZZYX23}
Ang Li, Bin Liang, Jingqian Zhao, Bowen Zhang, Min Yang, and Ruifeng Xu. 2023.
\newblock \href {https://aclanthology.org/2023.emnlp-main.972} {Stance detection on social media with background knowledge}.
\newblock In \emph{Proceedings of the 2023 Conference on Empirical Methods in Natural Language Processing, {EMNLP} 2023, Singapore, December 6-10, 2023}, pages 15703--15717. Association for Computational Linguistics.

\bibitem[{Li and Liang(2021)}]{li-liang-2021-prefix}
Xiang~Lisa Li and Percy Liang. 2021.
\newblock \href {https://doi.org/10.18653/v1/2021.acl-long.353} {Prefix-tuning: Optimizing continuous prompts for generation}.
\newblock In \emph{Proceedings of the 59th Annual Meeting of the Association for Computational Linguistics and the 11th International Joint Conference on Natural Language Processing (Volume 1: Long Papers)}, pages 4582--4597, Online. Association for Computational Linguistics.

\bibitem[{Li and Caragea(2023)}]{DBLP:conf/acl/LiC23}
Yingjie Li and Cornelia Caragea. 2023.
\newblock \href {https://doi.org/10.18653/V1/2023.FINDINGS-ACL.393} {Distilling calibrated knowledge for stance detection}.
\newblock In \emph{Findings of the Association for Computational Linguistics: {ACL} 2023, Toronto, Canada, July 9-14, 2023}, pages 6316--6329. Association for Computational Linguistics.

\bibitem[{Liang et~al.(2022{\natexlab{a}})Liang, Chen, Gui, He, Yang, and Xu}]{liang2022zero}
Bin Liang, Zixiao Chen, Lin Gui, Yulan He, Min Yang, and Ruifeng Xu. 2022{\natexlab{a}}.
\newblock Zero-shot stance detection via contrastive learning.
\newblock In \emph{Proceedings of the ACM Web Conference 2022}, pages 2738--2747.

\bibitem[{Liang et~al.(2022{\natexlab{b}})Liang, Zhu, Li, Yang, Gui, He, and Xu}]{liang2022jointcl}
Bin Liang, Qinglin Zhu, Xiang Li, Min Yang, Lin Gui, Yulan He, and Ruifeng Xu. 2022{\natexlab{b}}.
\newblock \href {https://doi.org/10.18653/v1/2022.acl-long.7} {{J}oint{CL}: A joint contrastive learning framework for zero-shot stance detection}.
\newblock In \emph{Proceedings of the 60th Annual Meeting of the Association for Computational Linguistics (Volume 1: Long Papers)}, pages 81--91, Dublin, Ireland. Association for Computational Linguistics.

\bibitem[{Liu et~al.(2021{\natexlab{a}})Liu, Yuan, Fu, Jiang, Hayashi, and Neubig}]{Liu-etal-2021-Pre-train}
Pengfei Liu, Weizhe Yuan, Jinlan Fu, Zhengbao Jiang, Hiroaki Hayashi, and Graham Neubig. 2021{\natexlab{a}}.
\newblock \href {https://arxiv.org/abs/2107.13586} {Pre-train, prompt, and predict: {A} systematic survey of prompting methods in natural language processing}.
\newblock \emph{ArXiv preprint}, abs/2107.13586.

\bibitem[{Liu et~al.(2023)Liu, Yuan, Fu, Jiang, Hayashi, and Neubig}]{DBLP:journals/csur/LiuYFJHN23}
Pengfei Liu, Weizhe Yuan, Jinlan Fu, Zhengbao Jiang, Hiroaki Hayashi, and Graham Neubig. 2023.
\newblock \href {https://doi.org/10.1145/3560815} {Pre-train, prompt, and predict: {A} systematic survey of prompting methods in natural language processing}.
\newblock \emph{{ACM} Comput. Surv.}, 55(9):195:1--195:35.

\bibitem[{Liu et~al.(2021{\natexlab{b}})Liu, Lin, Tan, and Wang}]{liu-etal-2021-enhancing}
Rui Liu, Zheng Lin, Yutong Tan, and Weiping Wang. 2021{\natexlab{b}}.
\newblock \href {https://doi.org/10.18653/v1/2021.findings-acl.278} {Enhancing zero-shot and few-shot stance detection with commonsense knowledge graph}.
\newblock In \emph{Findings of the Association for Computational Linguistics: ACL-IJCNLP 2021}, pages 3152--3157, Online. Association for Computational Linguistics.

\bibitem[{Liu et~al.(2021{\natexlab{c}})Liu, Ji, Fu, Du, Yang, and Tang}]{Liu-etal-2021-P-Tuning}
Xiao Liu, Kaixuan Ji, Yicheng Fu, Zhengxiao Du, Zhilin Yang, and Jie Tang. 2021{\natexlab{c}}.
\newblock \href {https://arxiv.org/abs/2110.07602} {P-tuning v2: Prompt tuning can be comparable to fine-tuning universally across scales and tasks}.
\newblock \emph{ArXiv preprint}, abs/2110.07602.

\bibitem[{Liu et~al.(2019{\natexlab{a}})Liu, Ott, Goyal, Du, Joshi, Chen, Levy, Lewis, Zettlemoyer, and Stoyanov}]{DBLP:journals/corr/abs-1907-11692}
Yinhan Liu, Myle Ott, Naman Goyal, Jingfei Du, Mandar Joshi, Danqi Chen, Omer Levy, Mike Lewis, Luke Zettlemoyer, and Veselin Stoyanov. 2019{\natexlab{a}}.
\newblock \href {http://arxiv.org/abs/1907.11692} {Roberta: {A} robustly optimized {BERT} pretraining approach}.
\newblock \emph{CoRR}, abs/1907.11692.

\bibitem[{Liu et~al.(2019{\natexlab{b}})Liu, Ott, Goyal, Du, Joshi, Chen, Levy, Lewis, Zettlemoyer, and Stoyanov}]{liu2019roberta}
Yinhan Liu, Myle Ott, Naman Goyal, Jingfei Du, Mandar Joshi, Danqi Chen, Omer Levy, Mike Lewis, Luke Zettlemoyer, and Veselin Stoyanov. 2019{\natexlab{b}}.
\newblock \href {https://arxiv.org/abs/1907.11692} {Roberta: A robustly optimized bert pretraining approach}.
\newblock \emph{ArXiv preprint}, abs/1907.11692.

\bibitem[{Liu et~al.(2021{\natexlab{d}})Liu, Lin, Cao, Hu, Wei, Zhang, Lin, and Guo}]{liu2021swin}
Ze~Liu, Yutong Lin, Yue Cao, Han Hu, Yixuan Wei, Zheng Zhang, Stephen Lin, and Baining Guo. 2021{\natexlab{d}}.
\newblock Swin transformer: Hierarchical vision transformer using shifted windows.
\newblock In \emph{Proceedings of the IEEE/CVF International Conference on Computer Vision}, pages 10012--10022.

\bibitem[{Maas et~al.(2013)Maas, Hannun, Ng et~al.}]{maas2013rectifier}
Andrew~L Maas, Awni~Y Hannun, Andrew~Y Ng, et~al. 2013.
\newblock Rectifier nonlinearities improve neural network acoustic models.
\newblock In \emph{Proc. icml}, volume~30, page~3. Atlanta, Georgia, USA.

\bibitem[{Mohammad et~al.(2016)Mohammad, Kiritchenko, Sobhani, Zhu, and Cherry}]{mohammad-etal-2016-semeval}
Saif Mohammad, Svetlana Kiritchenko, Parinaz Sobhani, Xiaodan Zhu, and Colin Cherry. 2016.
\newblock \href {https://doi.org/10.18653/v1/S16-1003} {{S}em{E}val-2016 task 6: Detecting stance in tweets}.
\newblock In \emph{Proceedings of the 10th International Workshop on Semantic Evaluation ({S}em{E}val-2016)}, pages 31--41, San Diego, California. Association for Computational Linguistics.

\bibitem[{Mutlu et~al.(2020)Mutlu, Oghaz, Jasser, Tutunculer, Rajabi, Tayebi, Ozmen, and Garibay}]{mutlu2020stance}
Ece~C Mutlu, Toktam Oghaz, Jasser Jasser, Ege Tutunculer, Amirarsalan Rajabi, Aida Tayebi, Ozlem Ozmen, and Ivan Garibay. 2020.
\newblock A stance data set on polarized conversations on twitter about the efficacy of hydroxychloroquine as a treatment for covid-19.
\newblock \emph{Data in brief}, 33:106401.

\bibitem[{Nguyen et~al.(2020)Nguyen, Vu, and Nguyen}]{DBLP:conf/emnlp/NguyenVN20}
Dat~Quoc Nguyen, Thanh Vu, and Anh~Tuan Nguyen. 2020.
\newblock \href {https://doi.org/10.18653/V1/2020.EMNLP-DEMOS.2} {Bertweet: {A} pre-trained language model for english tweets}.
\newblock In \emph{Proceedings of the 2020 Conference on Empirical Methods in Natural Language Processing: System Demonstrations, {EMNLP} 2020 - Demos, Online, November 16-20, 2020}, pages 9--14. Association for Computational Linguistics.

\bibitem[{Radford et~al.(2021)Radford, Kim, Hallacy, Ramesh, Goh, Agarwal, Sastry, Askell, Mishkin, Clark, Krueger, and Sutskever}]{radford2021learning}
Alec Radford, Jong~Wook Kim, Chris Hallacy, Aditya Ramesh, Gabriel Goh, Sandhini Agarwal, Girish Sastry, Amanda Askell, Pamela Mishkin, Jack Clark, Gretchen Krueger, and Ilya Sutskever. 2021.
\newblock \href {http://proceedings.mlr.press/v139/radford21a.html} {Learning transferable visual models from natural language supervision}.
\newblock In \emph{Proceedings of the 38th International Conference on Machine Learning, {ICML} 2021, 18-24 July 2021, Virtual Event}, volume 139 of \emph{Proceedings of Machine Learning Research}, pages 8748--8763. {PMLR}.

\bibitem[{Shin et~al.(2020)Shin, Razeghi, Logan~IV, Wallace, and Singh}]{shin-etal-2020-AutoPrompt}
Taylor Shin, Yasaman Razeghi, Robert~L. Logan~IV, Eric Wallace, and Sameer Singh. 2020.
\newblock \href {https://doi.org/10.18653/v1/2020.emnlp-main.346} {{A}uto{P}rompt: {E}liciting {K}nowledge from {L}anguage {M}odels with {A}utomatically {G}enerated {P}rompts}.
\newblock In \emph{Proceedings of the 2020 Conference on Empirical Methods in Natural Language Processing (EMNLP)}, pages 4222--4235, Online. Association for Computational Linguistics.

\bibitem[{Somasundaran and Wiebe(2010)}]{somasundaran2010recognizing}
Swapna Somasundaran and Janyce Wiebe. 2010.
\newblock \href {https://aclanthology.org/W10-0214} {Recognizing stances in ideological on-line debates}.
\newblock In \emph{Proceedings of the {NAACL} {HLT} 2010 Workshop on Computational Approaches to Analysis and Generation of Emotion in Text}, pages 116--124, Los Angeles, CA. Association for Computational Linguistics.

\bibitem[{Sun et~al.(2018)Sun, Wang, Zhu, and Zhou}]{sun-etal-2018-stance}
Qingying Sun, Zhongqing Wang, Qiaoming Zhu, and Guodong Zhou. 2018.
\newblock \href {https://aclanthology.org/C18-1203} {Stance detection with hierarchical attention network}.
\newblock In \emph{Proceedings of the 27th International Conference on Computational Linguistics}, pages 2399--2409, Santa Fe, New Mexico, USA. Association for Computational Linguistics.

\bibitem[{Touvron et~al.(2023)Touvron, Martin, Stone, Albert, Almahairi, Babaei, Bashlykov, Batra, Bhargava, Bhosale, Bikel, Blecher, Canton{-}Ferrer, Chen, Cucurull, Esiobu, Fernandes, Fu, Fu, Fuller, Gao, Goswami, Goyal, Hartshorn, Hosseini, Hou, Inan, Kardas, Kerkez, Khabsa, Kloumann, Korenev, Koura, Lachaux, Lavril, Lee, Liskovich, Lu, Mao, Martinet, Mihaylov, Mishra, Molybog, Nie, Poulton, Reizenstein, Rungta, Saladi, Schelten, Silva, Smith, Subramanian, Tan, Tang, Taylor, Williams, Kuan, Xu, Yan, Zarov, Zhang, Fan, Kambadur, Narang, Rodriguez, Stojnic, Edunov, and Scialom}]{DBLP:journals/corr/abs-2307-09288}
Hugo Touvron, Louis Martin, Kevin Stone, Peter Albert, Amjad Almahairi, Yasmine Babaei, Nikolay Bashlykov, Soumya Batra, Prajjwal Bhargava, Shruti Bhosale, Dan Bikel, Lukas Blecher, Cristian Canton{-}Ferrer, Moya Chen, Guillem Cucurull, David Esiobu, Jude Fernandes, Jeremy Fu, Wenyin Fu, Brian Fuller, Cynthia Gao, Vedanuj Goswami, Naman Goyal, Anthony Hartshorn, Saghar Hosseini, Rui Hou, Hakan Inan, Marcin Kardas, Viktor Kerkez, Madian Khabsa, Isabel Kloumann, Artem Korenev, Punit~Singh Koura, Marie{-}Anne Lachaux, Thibaut Lavril, Jenya Lee, Diana Liskovich, Yinghai Lu, Yuning Mao, Xavier Martinet, Todor Mihaylov, Pushkar Mishra, Igor Molybog, Yixin Nie, Andrew Poulton, Jeremy Reizenstein, Rashi Rungta, Kalyan Saladi, Alan Schelten, Ruan Silva, Eric~Michael Smith, Ranjan Subramanian, Xiaoqing~Ellen Tan, Binh Tang, Ross Taylor, Adina Williams, Jian~Xiang Kuan, Puxin Xu, Zheng Yan, Iliyan Zarov, Yuchen Zhang, Angela Fan, Melanie Kambadur, Sharan Narang, Aur{\'{e}}lien Rodriguez, Robert Stojnic, Sergey Edunov,
  and Thomas Scialom. 2023.
\newblock \href {https://doi.org/10.48550/ARXIV.2307.09288} {Llama 2: Open foundation and fine-tuned chat models}.
\newblock \emph{CoRR}, abs/2307.09288.

\bibitem[{Velickovic et~al.(2018)Velickovic, Cucurull, Casanova, Romero, Li{\`{o}}, and Bengio}]{veličković2018graph}
Petar Velickovic, Guillem Cucurull, Arantxa Casanova, Adriana Romero, Pietro Li{\`{o}}, and Yoshua Bengio. 2018.
\newblock \href {https://openreview.net/forum?id=rJXMpikCZ} {Graph attention networks}.
\newblock In \emph{6th International Conference on Learning Representations, {ICLR} 2018, Vancouver, BC, Canada, April 30 - May 3, 2018, Conference Track Proceedings}. OpenReview.net.

\bibitem[{Wei et~al.(2020)Wei, Zhang, Li, Zhang, and Wu}]{wei2020multi}
Xi~Wei, Tianzhu Zhang, Yan Li, Yongdong Zhang, and Feng Wu. 2020.
\newblock \href {https://doi.org/10.1109/CVPR42600.2020.01095} {Multi-modality cross attention network for image and sentence matching}.
\newblock In \emph{2020 {IEEE/CVF} Conference on Computer Vision and Pattern Recognition, {CVPR} 2020, Seattle, WA, USA, June 13-19, 2020}, pages 10938--10947. {IEEE}.

\bibitem[{Weinzierl and Harabagiu(2023)}]{weinzierl-harabagiu-2023-identification}
Maxwell Weinzierl and Sanda Harabagiu. 2023.
\newblock \href {https://doi.org/10.18653/v1/2023.emnlp-main.776} {Identification of multimodal stance towards frames of communication}.
\newblock In \emph{Proceedings of the 2023 Conference on Empirical Methods in Natural Language Processing}, pages 12597--12609, Singapore. Association for Computational Linguistics.

\bibitem[{Wen and Hauptmann(2023)}]{wen-hauptmann-2023-zero}
Haoyang Wen and Alexander Hauptmann. 2023.
\newblock \href {https://doi.org/10.18653/v1/2023.acl-short.127} {Zero-shot and few-shot stance detection on varied topics via conditional generation}.
\newblock In \emph{Proceedings of the 61st Annual Meeting of the Association for Computational Linguistics (Volume 2: Short Papers)}, pages 1491--1499, Toronto, Canada. Association for Computational Linguistics.

\bibitem[{Zhang et~al.(2020)Zhang, Yang, Li, Ye, Xu, and Dai}]{zhang-etal-2020-enhancing-cross}
Bowen Zhang, Min Yang, Xutao Li, Yunming Ye, Xiaofei Xu, and Kuai Dai. 2020.
\newblock \href {https://doi.org/10.18653/v1/2020.acl-main.291} {Enhancing cross-target stance detection with transferable semantic-emotion knowledge}.
\newblock In \emph{Proceedings of the 58th Annual Meeting of the Association for Computational Linguistics}, pages 3188--3197, Online. Association for Computational Linguistics.

\bibitem[{Zhao et~al.(2023)Zhao, Li, and Caragea}]{DBLP:conf/acl/ZhaoLC23}
Chenye Zhao, Yingjie Li, and Cornelia Caragea. 2023.
\newblock \href {https://doi.org/10.18653/V1/2023.ACL-LONG.747} {{C-STANCE:} {A} large dataset for chinese zero-shot stance detection}.
\newblock In \emph{Proceedings of the 61st Annual Meeting of the Association for Computational Linguistics (Volume 1: Long Papers), {ACL} 2023, Toronto, Canada, July 9-14, 2023}, pages 13369--13385. Association for Computational Linguistics.

\bibitem[{Zheng et~al.(2022)Zheng, Sun, Yang, and Xu}]{DBLP:conf/emnlp/ZhengSYX22}
Kai Zheng, Qingfeng Sun, Yaming Yang, and Fei Xu. 2022.
\newblock \href {https://doi.org/10.18653/V1/2022.FINDINGS-EMNLP.83} {Knowledge stimulated contrastive prompting for low-resource stance detection}.
\newblock In \emph{Findings of the Association for Computational Linguistics: {EMNLP} 2022, Abu Dhabi, United Arab Emirates, December 7-11, 2022}, pages 1168--1178. Association for Computational Linguistics.

\bibitem[{Zhou et~al.(2022)Zhou, Yang, Loy, and Liu}]{zhou-etal-2022-Learning}
Kaiyang Zhou, Jingkang Yang, Chen~Change Loy, and Ziwei Liu. 2022.
\newblock \href {https://doi.org/10.1007/s11263-022-01653-1} {Learning to prompt for vision-language models}.
\newblock \emph{Int. J. Comput. Vis.}, 130(9):2337--2348.

\end{thebibliography}


\appendix

\section{Keywords}
\label{sec:keywords}
In this section, we introduce the keywords to retrieve tweets.

\subsection{Multi-modal Twitter Stance Election 2020}

\begin{itemize}
\setlength\itemsep{3pt}
    \item Since Twitter Stance Election 2020~\cite{kawintiranon-singh-2021-knowledge} didn't explicitly give the Keywords for collecting tweets, We used the keywords related to the election while with no clear preference: one of \textit{\#vote, \#Debates2020, \#USElection2020, \#PresidentialDebate2020, \#2020Election, \#votersuppression, \#GetOuttheVote, \#2020elections} + mention of Trump or Biden.
    \item Filter for posting time: \\
    \vspace{-7pt} \\
    \begin{tabular}{lc}
    DT   & 01/01/2020 $\rightarrow$ {09/30/2020}\\
    JB   & 01/01/2020 $\rightarrow$ {09/30/2020}
    \end{tabular}
\end{itemize}

\subsection{Multi-modal COVID-CQ}

\begin{itemize}
\setlength\itemsep{3pt}
    \item We followed the keywords for collecting tweets from COVID-CQ~\cite{mutlu2020stance}: one of \textit{hydroxychloroquine, chloroquine, HCQ}.
    \item Filter for posting time: \\
    \vspace{-7pt} \\
    \begin{tabular}{lc}
    CQ   & 04/01/2020 $\rightarrow$ {04/30/2020}
    \end{tabular}
\end{itemize}

\subsection{Multi-modal Will-They-Won’t-They}

\begin{itemize}
\setlength\itemsep{3pt}
    \item We followed the keywords for collecting tweets from Will-They-Won’t-They~\cite{conforti-etal-2020-will}: one of \textit{merge, acquisition, agreement, acquire, takeover, buyout, integration} + mention of a given company/acronym.
    \item Filter for posting time: \\
    \vspace{-7pt} \\
    \begin{tabular}{lc}
    CVS\_AET   & 02/15/2017 $\rightarrow$ {12/17/2018}\\
    CI\_ESRX   & 05/27/2017 $\rightarrow$ {09/17/2018}\\
    ANTM\_CI   & 04/01/2014 $\rightarrow$ {04/28/2017}\\
    AET\_HUM   & 09/01/2014 $\rightarrow$ {01/23/2017}\\
    DIS\_FOXA  & 07/09/2017 $\rightarrow$ {04/18/2018}
    \end{tabular}
\end{itemize}

\subsection{Multi-modal Russo-Ukrainian Conflict}

\begin{itemize}
\setlength\itemsep{3pt}
    \item We used the keywords: \textit{Ukraine, Russia, Putin, Zelensky, Ukrainian, Russian}
    \item Filter for posting time: \\
    \vspace{-7pt} \\
    \begin{tabular}{lc}
    RUS   & 01/01/2022 $\rightarrow$ {06/30/2023}\\
    UKR   & 01/01/2022 $\rightarrow$ {06/30/2023}
    \end{tabular}
\end{itemize}

\subsection{Multi-modal Taiwan Question}

\begin{itemize}
\setlength\itemsep{3pt}
    \item We used the keywords: \textit{Taiwan, Taiwan Crisis, Nancy Pelosi, Taiwan Strait}
    \item Filter for posting time: \\
    \vspace{-7pt} \\
    \begin{tabular}{lc}
    MOC   & 01/01/2022 $\rightarrow$ {06/30/2023}\\
    TOC   & 01/01/2022 $\rightarrow$ {06/30/2023}
    \end{tabular}
\end{itemize}

\section{Annotation Guidelines}
\label{sec:guidelines}
To ensure consistency with previous stance detection work, we follow the guidelines of Twitter Stance Election 2020~\cite{kawintiranon-singh-2021-knowledge}, COVID-CQ \cite{mutlu2020stance}, and Will-They-Won't-They \cite{conforti-etal-2020-will} to annotate the multi-modal stance of \textsc{Mtse}, \textsc{Mccq} and \textsc{Mwtwt}. For \textsc{Mruc} and \textsc{Mtwq}, the annotation guidelines are shown below:

\subsection{Annotation Guidelines of MRUC and MTWQ}
The annotation process consists of choosing one of three possible labels, given a tweet and an image. The three labels to choose from are Support, Oppose and Neutral.

\textbf{Label 1: Support -} If the tweet and image use direct or indirect expressions to support the target, or support objects which can represent the target (such as leaders, events).

\textbf{Label 1: Oppose -} If the tweet and image use direct or indirect expressions to oppose the target, or oppose objects which can represent the target (such as leaders, events).

\textbf{Label 1: Neutral -} If the tweet and image only mention the target without expressing a stance.

\subsection{Details of Datasets Partition}
As mentioned in Section~\ref{subsec:data_partition}, for each task/target, the dataset is divided into a training set, a development set, and a testing set with a ratio of 70\%:10\%:20\%. In order to ensure that the partition is not affected by the data distribution bias. We performed 20 different random divisions for each task using the ratios above. We use the two baselines: BERT and BERT+ViT, to test every 20 groups of divisions. For each task, we take the division that can make the results of two baseline close to the median as the final partition.

\section{Prompt Tuning Analysis}
\label{prompt-tuning-analysis}

\subsection{Analysis of Textual Prompt Tuning}
\label{textual-prompt}

\paragraph{Frozen vs Tuned}For textual prompt tuning, we utilize a frozen paradigm to make full use of the semantic information learned by the pre-trained language model. To analyze the effectiveness of the frozen paradigm, we also tried a tuned soft prompt, the results are shown in Table~\ref{results-fix-prompt}. The results of the tuned paradigm are extremely poorer than the frozen one and fluctuate greatly. One possible reason is that, unlike the continuous picture patch in the visual transformer, the token in the textual modality is discrete single words, so it is hard to use a gradient-based method to fine-tune the soft prompts like a visual modality. Thus, in this paper, we choose a manually designed fixed prompt to obtain better performance.

\begin{table}[!t]
\small
\centering
\setlength{\tabcolsep}{2.5pt}
\renewcommand{\arraystretch}{1.1}
\begin{tabular}{l|ccccc}
\hline
 {\textsc{Method}} & \textsc{Mtse} & \textsc{Mccq} & \textsc{Mwtwt} & \textsc{Mruc} & \textsc{Mtwq} \\
 \hline
\cellcolor[gray]{0.92}Frozen & \cellcolor[gray]{0.92}{\bf 60.84} & \cellcolor[gray]{0.92}{\bf 67.67} & \cellcolor[gray]{0.92}{\bf 77.59} & \cellcolor[gray]{0.92}{\bf 75.76} & \cellcolor[gray]{0.92}{\bf 67.59} \\
\cdashline{1-4}[2pt/3pt]
Tuned & 53.85 & 58.41 & 62.58 & 59.64 & 52.26 \\
 \hline
\end{tabular}
\caption{Experimental results of fixed prompt and tuned soft prompt of textual modality. The reported results are the Macro F1-score across all targets in a dataset on in-target multi-modal stance detection.}
\label{results-fix-prompt}
\end{table}

\paragraph{Hand-design textual prompts}To analyze the impact of the type of the hand-design textual prompts, we design several types of textual prompts and report the experiments in Table~\ref{results-textual-prompts}. Take the \textsc{Mtse} dataset as an example, it can be seen that the experimental results of all types of prompts are superior to those without prompts (BERT+ViT). This demonstrates the significance of the targeted prompts in this task. Further, we can also see that there are considerable differences in performance between different types of prompts. Specifically, when using both target and stance to design prompts (\textcircled{3}, \textcircled{4}, and \textcircled{5}), the model performs significantly better. Therefore, in our method, we choose the type of \textcircled{5} to devise the textual prompts for multi-modal stance detection.

\begin{table}[!t]
\small
\centering
\setlength{\tabcolsep}{2.5pt}
\renewcommand{\arraystretch}{1.1}
\begin{tabular}{l|p{0.55\linewidth}|c}
\hline
 {\textsc{Method}} & \multicolumn{1}{c|}{Textual Prompts} & \textsc{Mtse} \\
 \hline
BERT+ViT & - & 53.29 \\
  \hline
\textcircled{1} & Trump & 58.54  \\
\textcircled{2} & Donald Trump & 59.53  \\
\textcircled{3} & stance on Donald Trump & 60.24  \\
\multirow{2}*{\textcircled{4}} & What is the stance on Donald Trump? & 58.85  \\
\cdashline{1-3}[2pt/3pt]
\cellcolor[gray]{0.92}\textcircled{5} (Ours) & \cellcolor[gray]{0.92}{The stance on Donald Trump is:} & \cellcolor[gray]{0.92}{\bf 60.84} \\
 \hline
\end{tabular}
\caption{Experimental results of using different textual prompts in MTSE dataset on target ``Donald Trump''.}
\label{results-textual-prompts}
\end{table}

\subsection{Analysis of Visual Prompt Tuning}
\label{ana-visual-prompting}

\paragraph{Depth of prompt tuning}Following \cite{jia2022vpt}, we conduct experiments with different depths of visual prompt tuning to analyze the impact of the depth of prompt tuning in our model.
Here, shallow prompt tuning refers to only fine-tuning the prompt tokens in the Embedding layer of ViT, while deep prompt tuning refers to fine-tuning the prompt tokens in the Embedding layer of ViT and each layer in the transformer. 
The experimental results are shown in Table~\ref{results-shallow}. We can see that shallow prompt tuning has a better overall effect.

\begin{table}[!t]
\small
\centering
\setlength{\tabcolsep}{2.5pt}
\renewcommand{\arraystretch}{1.1}
\begin{tabular}{l|ccccc}
\hline
 {\textsc{Method}} & \textsc{Mtse} & \textsc{Mccq} & \textsc{Mwtwt} & \textsc{Mruc} & \textsc{Mtwq}\\
 \hline
\cellcolor[gray]{0.92}Shallow (Ours) & \cellcolor[gray]{0.92}{60.84} & \cellcolor[gray]{0.92}{\bf 67.67} & \cellcolor[gray]{0.92}{\bf 77.59} & \cellcolor[gray]{0.92}{\bf 75.76} & \cellcolor[gray]{0.92}{\bf 67.59} \\
\cdashline{1-6}[2pt/3pt]
Deep & \textbf{63.48} & 62.10 & 72.68 & 69.66 & 61.08 \\
 \hline
\end{tabular}
\caption{Comparison results of using the shallow prompt tuning and deep prompt tuning. Shallow prompt tuning refers to only fine-tuning the prompt tokens in the Embedding layer of ViT, while deep prompt tuning refers to fine-tuning the prompt tokens in the Embedding layer of ViT and each layer in the Transformer. The reported results are the Macro F1-score across all targets in a dataset on in-target multi-modal stance detection.}
\label{results-shallow}
\end{table}

\begin{figure}[!t]
\centering  
\includegraphics[width=1\linewidth]{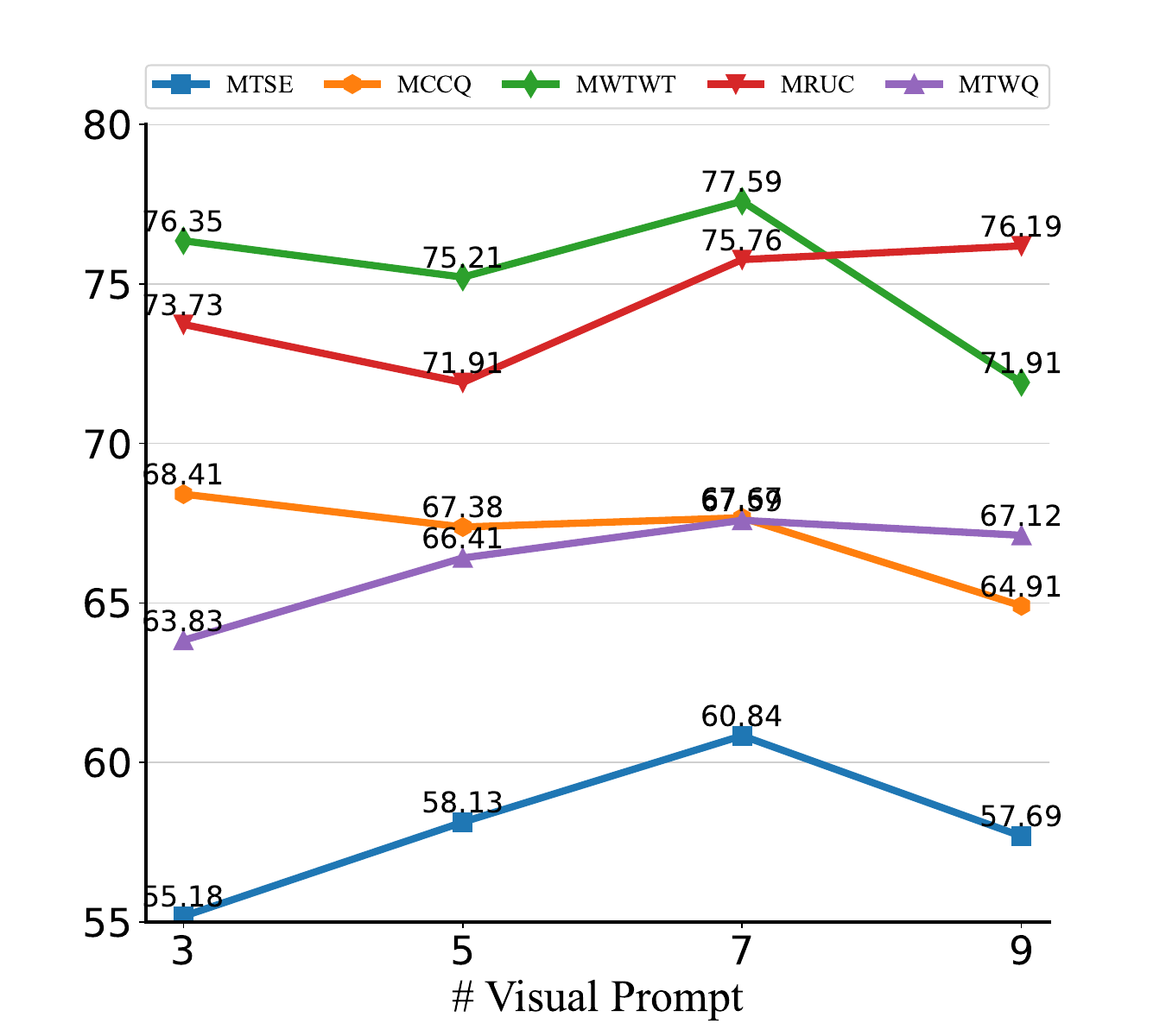}
\caption{Experimental results of using different numbers of visual prompting tuning tokens. The reported results are the Macro F1-score across all targets in a dataset on in-target multi-modal stance detection.}
\label{fig-prompt-tokens}
\end{figure}

\paragraph{Number of visual prompt tokens}To analyze the impact of the number of visual prompt tokens, we set the value range of tokens as \{3, 5, 7, 9\} for comparative experiments. The results are shown in Figure~\ref{fig-prompt-tokens}. Note that different values of tokens can have a certain impact on performance. When the value is 7, the overall performance of the model on all datasets is the best. Therefore, we set the number of tokens to 7 in our method.

\end{document}